\documentclass[review]{elsarticle}

\journal{Information Sciences}

\usepackage{amssymb}
\usepackage{graphicx}
\usepackage[table,xcdraw]{xcolor}
\usepackage{subcaption,booktabs}
\usepackage{mathtools}
\usepackage{rotating}
\usepackage{afterpage}
\usepackage{adjustbox}
\usepackage{longtable}
\usepackage{multirow}
\usepackage{lscape}
\usepackage{hhline}
\usepackage{verbatim}
\newcommand*\rot{\rotatebox{90}}

\usepackage{lineno}

\newdefinition{rmk}{Remark}

\begin{document}

\begin{frontmatter}




\title{An Analysis of the Admissibility of the Objective Functions Applied in Evolutionary Multi-objective Clustering}


\author[add1] {Cristina Y. Morimoto\corref{mycorrespondingauthor}}
\ead{cristina.morimoto@ufpr.br}

\author[add1] {Aurora Pozo}
\ead{aurora@inf.ufpr.br}

\author[add2] {Marc\'ilio C. P. de Souto}
\ead{marcilio.desouto@univ-orleans.fr}

\address[add1]{Federal University of Paran\'a, Curitiba-PR, Brazil}
\address[add2]{LIFO/University of Orl\'eans, Orl\'eans, France}
\cortext[mycorrespondingauthor]{Corresponding author}

\begin{abstract}

A variety of clustering criteria has been applied as an objective function in Evolutionary Multi-Objective Clustering approaches (EMOCs). However, most EMOCs do not provide detailed analysis regarding the choice and usage of the objective functions. Aiming to support a better choice and definition of the objectives in the EMOCs, this paper proposes an analysis of the admissibility of the clustering criteria in evolutionary optimization by examining the search direction and its potential in finding optimal results. As a result, we demonstrate how the admissibility of the objective functions can influence the optimization. Furthermore, we provide insights regarding the combinations and usage of the clustering criteria in the EMOCs.

\end{abstract}

\begin{keyword}
Clustering criteria \sep Multi-objective clustering  \sep Evolutionary multi-objective optimization \sep Clustering analysis.


\end{keyword}

\end{frontmatter}

\section{Introduction}
The use of knowledge discovery techniques has become essential to analyze and understand a large volume of data generated in different fields of application (e.g. marketing, medicine, bioinformatics). Clustering analysis has been widely studied and adopted for several purposes, including pattern analysis, image segmentation, data mining, and decision-making. 
Clustering is a type of unsupervised learning whose goal is to find the underlying structures that compose finite sets of data (clusters), in which the objects belonging to a cluster should share some relevant property (similarity) regarding the data domain~\cite{aggarwal2014data}. 

In recent years, multi-objective evolutionary algorithms (MOEAs) have become one popular methodology for clustering. However, the design and definition of the clustering problem are still challenges. In particular, issues concerning the definition of the objective functions emerge in evolutionary multi-objective optimization, in addition to other difficulties. For example, there are a variety of clustering criteria being applied in the Evolutionary Multi-Objective Clustering approaches (EMOCs), but most studies do not provide any analysis regarding the choice and combination of the objective functions, in which, fundamentally, the objective functions are selected concerning some desired proprieties in the data clustering. Thus, in this paper, we propose the analysis of the admissibility of the clustering criteria to support the  definition of the objective functions in the EMOCs. Here, we characterize admissible objective functions as having the property of detecting the ``natural'' ground-truth clustering, in which the ground-truth clustering has an optimal value.
In other words, our study introduces an investigation for the admissibility of the objective functions applied to evolutionary multi-objective clustering that supports defining whether the objectives are worth optimizing. 
It is important to note that we evaluate the admissibility regarding the search direction. This is different from~\cite{fisher1971admissible}, where the authors consider the admissibility of the clustering algorithms by considering the evaluation of the structure of the data essentially.

Other studies have evaluated the objective functions for evolutionary multi-objective data clustering. 
\cite{handl2012clustering} presents a comparison of four criteria pairs for multi-objective clustering in datasets with different types of clusters. \cite{Barton2015} investigated some clustering criteria and analyzed their correlation with the ground truth to develop an evolutionary multi-objective clustering algorithm. 
However, they do not consider evaluating the admissibility of the objective functions and the impact of the initial population in the evolutionary optimization.  
 
In particular, this paper presents the analysis of 17 objective functions as for their admissibility in 24 artificial datasets with different data structures. 
We used the objective functions present in the EMOCs found in the literature to analyze whether they can lead the EMOCs to optimal results. 
Hence, we analyzed how some initialization methods can affect evolutionary optimization. Furthermore, we correlate the admissibility analysis regarding the clustering results by considering some optimization scenarios with  different combinations of objective functions. As a result, our analysis provides insights regarding the choice of the objective functions and the initialization strategy to improve designing EMOCs. 
The remainder of this paper is organized as follows. In Section~\ref{section:background}, we present the main concepts of evolutionary multi-objective clustering and the clustering criteria used in our experiments.  Section~\ref{sec:experiment} describes the datasets considered, the specific configuration and settings of the compared methods, and the performance assessment adopted. Then, in Sections~\ref{sec:results} and~\ref{sec:disc}, we present and discuss the results of our experimental evaluation of the objective functions. Finally, Section~\ref{sec:remarks} highlights our main findings and discusses future works.

\section{Background}\label{section:background}
\subsection{Preliminaries}\label{section:definitions}
The convention adopted in this paper considers the terms: clustering criteria, objective functions, fitness functions, and heuristic functions interchangeably to represent the multi-objective clustering problem's goals or objectives (i.e., distinct mathematical functions) to be achieved:   
\begin{itemize}
\item \textbf{\textit{Clustering Criteria}}: A clustering criterion (function) guides the selection of features and clustering schemes in a clustering algorithm. 
\noindent
\item \textbf{\textit{Objective Functions}}: An objective function indicates how much each variable contributes to the value to be optimized in the problem. The objective function refers to a criterion that should be maximized or minimized in an optimization problem.
\noindent
\item \textbf{\textit{Fitness Functions}}: A fitness function is a particular type of objective function that quantifies the optimality of a solution.
The fitness functions are used in evolutionary approaches to guide the search towards optimal design solutions. 
\noindent
\item\textbf{\textit{Heuristic Functions}}: A heuristic is a term adopted in artificial intelligence (AI) that works by guiding search, suggesting behavior, making decisions, or transforming the problem. A heuristic function guides the decision, as a strategy or simplification, to limit the search for solutions in large problem spaces. However, they do not guarantee optimal solutions.
\end{itemize}

In this study, instead of using the term fitness function, we rely on the general term used in evolutionary multi-objective optimization: the objective function. However, each term is important to facilitate the general understanding of the content of this paper, considering that they relate to different fields of study. 

The term \textbf{optimal value} in clustering refers to \textbf{ground-truth} clustering, called \textbf{true partition}.
In other words, the true partition represents the ideal model partition. As our analysis considered artificial datasets, the true partition was known in advance, making it possible to perform a detailed examination of  the underlying structure of the data and relate it to the clustering criteria.

Other important terms applied in our study are regarding multi-objective optimization. It is essential to define the general aspects of objective functions and restrictions to understand the use of admissible or inadmissible objective functions in EMOCs.  
The \textbf{Multi-objective Optimization Problem (MOP)} can be defined as the optimization of the vector function $\mathbf{F}(\pi)$, Eq.~\ref{mop}, that maps a tuple of decision variables ($\pi$) to a tuple of $z$ objectives ($z \geq 2$) and satisfies inequality and equality restrictions (or constraints), Eq.~\ref{mop_r1}. These restrictions are used to  determine the feasible region of solutions~\cite{Zitzler1999}. 
\begin{equation}\label{mop}
\textrm{minimize/maximize } \mathbf{F}(\mathbf{\pi}) = (f_1(\mathbf{\pi}), f_2(\mathbf{\pi}), \ldots, f_{z}(\mathbf{\pi}))
\end{equation}
\begin{equation}\label{mop_r1}
\begin{split}
\textrm{subjected to }& g_i(\mathbf{\pi}) \le 0, \quad i=\{1, \ldots, p\}, \textrm{and } \\ &  h_j(\mathbf{\pi})=0, \quad j=\{1,  \ldots, q\} 
\end{split}
\end{equation}

Most multi-objective algorithms select solutions using the \textbf{Pareto dominance} relation. Formally, the Pareto dominance is define as: Let $\mathbf{\pi}_1$ and $\mathbf{\pi}_2$ be two feasible solutions; $\mathbf{\pi}_1$ is said to dominate $\mathbf{\pi}_2$ (denoted as $ \mathbf{\pi}_1 	\prec \mathbf{\pi}_2$), if the following conditions are satisfied  \cite{Coello2006}:
(i) $\forall i \in  \{1, 2, \ldots, z\}: f_i(\mathbf{\pi}_1) \leq f_i(\mathbf{\pi}_2)$ and (ii) $\exists j \in  \{1, 2, \ldots, z\}: f_j(\mathbf{\pi}_1)<f_j\mathbf{(\pi}_2)$. In particular, identifying a solution $\pi$ that is feasible and optimizes the objective at hand is notably challenging when restrictions and objectives have a non-linear, non-convex, discrete, or non-differentiable nature. One common approach to dealing with the restrictions is to treat those restrictions as objective functions. For example, in bi-objective optimization, a constraint can be used as a second objective subjected to multi-objective optimization for the formation of a Pareto front, in which the optimization can be focused on the main objective function.

\subsection{Admissible Heuristic Functions}\label{sec:adm}
In general, \textbf{an admissible heuristic function can be characterized as a function that does not overestimate the cost of reaching the goal}. 
The evaluation of a heuristic function considers the states in the search space, in which the cost of the actual state should be lower than or equal to the estimated cost of finding the optimal solution. Formally, a heuristic $h(s)$ is admissible if, for every state $s$, $h(s) \leq h^*(s)$, where $h^*(s)$ is the true cost to reach the goal state from $s$ \cite{russell2002artificial}.  

In our study, we consider this general admissibility concept in evolutionary optimization. Thus, we verified the potential of the objective functions, as heuristic functions, in finding the optimal results considering the search direction. Here, as above-mentioned, we considered the optimal results as the underlying structure of the data (true partition or ground-truth).
It is important to note that our study considered artificial datasets in which the optimal clustering results were already known. That made it possible to analyze the admissibility of the objective functions in the clustering problem.

\begin{figure}[!htb]
	\centering
	\includegraphics[scale=0.45]{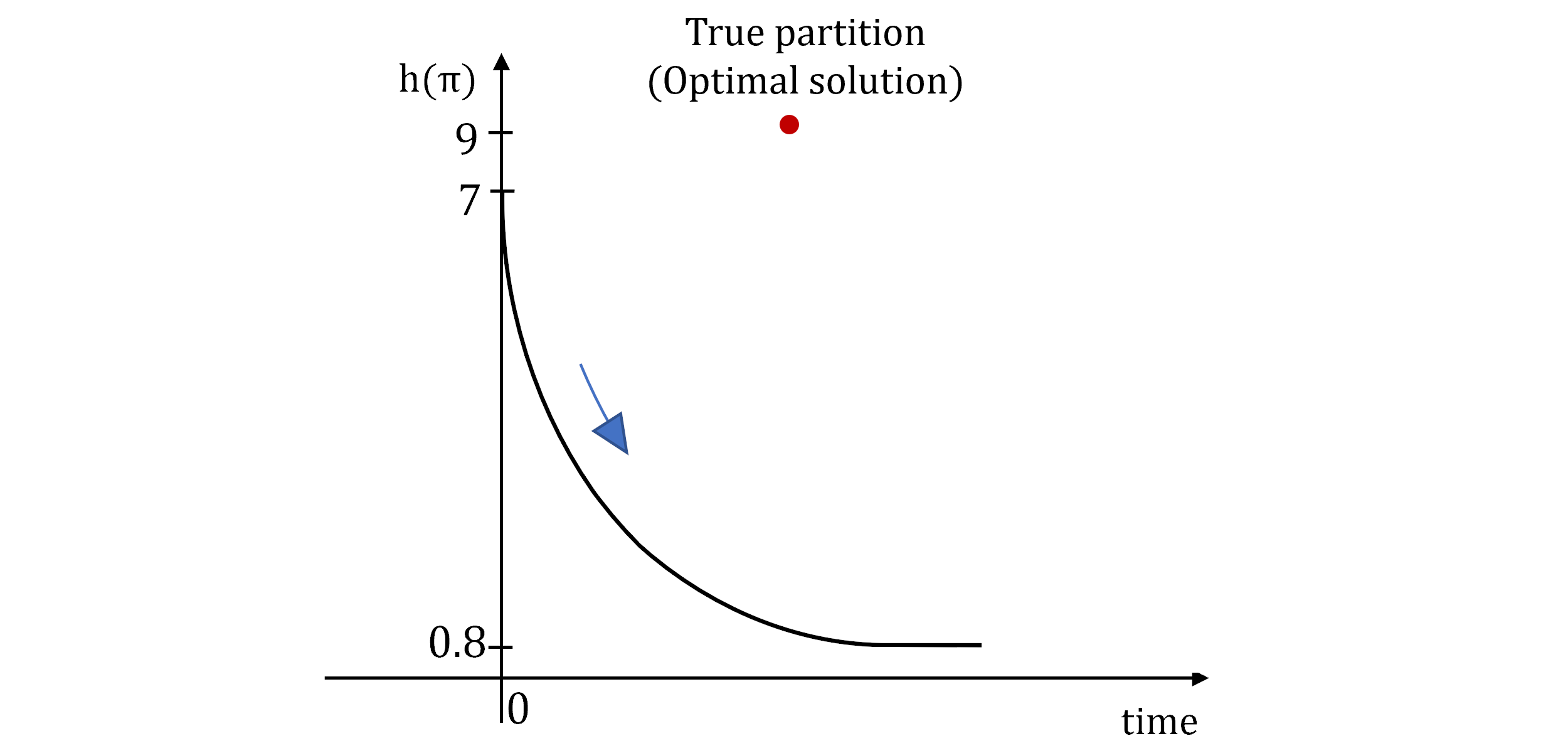}\caption{Example of an inadmissible $h(\pi)$}\label{fig:adm}
\end{figure}

In practice, our analysis consists of evaluating the inadmissibility of the objective functions. A heuristic is inadmissible if $h(s) \geq h^*(s)$ for the maximization problem, and $h(s) \leq h^*(s)$ for a minimization problem. 
Fig. \ref{fig:adm} illustrates the general case of the inadmissibility applied in our analysis, in which the red point represents the optimal solution, and the blue arrow indicates the search direction. Assuming a minimization problem, $h(\mathbf{\pi})$ is inadmissible in order to find the optimal solution, in which $h(\mathbf{\pi})=7$ at time 0, and $h^*(\mathbf{\pi})=9$. Furthermore, the optimization of the $h(\mathbf{\pi})$ worsens this aspect. 

\textbf{Inadmissible functions are not adequate to be optimized; however, they could be applied as restrictions} (Eq.~\ref{mop_r1}) to define the feasible region of solutions. In particular, the inadmissible functions can be applied as objective functions to constrain the search in a specific direction. 

\subsection{Objective Functions}\label{sec:obj}

In this section, we present the objective functions used in our study: four compactness criteria (intra-cluster entropy, overall deviation, intra-cluster variance, and total within-cluster variance), two connectedness criteria (connectivity and data continuity degree), four separation criteria (average between-group sum of squares, average separation, separation index, and graph-based separation), and seven compactness and separation criteria (Calinski-Harabasz, Davies-Bouldin, Dunn, modularity, silhouette, Pakhira-Bandyopadhyay-Maulik, and Xeni-Beny). These objective functions were extracted from the evolutionary multi-objective clustering approaches detailed in Section \ref{sec:emoc}.

We used the following notation in the equations presented in this section: $n$ refers to the number of objects in the dataset $\mathbf{X}$, $\pi$  denotes a partition, $k$ denotes the number of clusters in $\pi$,  $\mathbf{c}_i$ refers to the  $i$th cluster that belongs to $\pi$, $\mathbf{x}_a$ denotes a generic object, $n_i$ denotes the number of objects in $\mathbf{c}_i$, $\mathbf{z}_i$  refers to the centroid of cluster $\mathbf{c}_i$, and $\overline{\mathbf{z}}$ represents the centroid of the dataset. Furthermore, $d(.,.)$ denotes the chosen distance function (Euclidean distance).

\subsubsection{Intra-cluster Entropy} 
The {intra-cluster entropy} ($Ent$) measures the degree of similarity between each cluster center and the data objects that belong to that cluster, as the probability of grouping all the data objects into that particular cluster. A larger value  (maximization) of this index implies better clustering~\cite{Ripon2006_IJCNN}. The $Ent$ is defined in Eq.~\ref{entropy}, in which $g(\mathbf{z}_i)$ is the average similarity between $\mathbf{z}_i$ and the data object belong to cluster $\mathbf{c}_i$, and the $cos(.,.)$ denote the cosine distance.
\begin{equation}\label{entropy}
\begin{gathered}
Ent(\pi) = \sum^k_{i=1} \left [(1-H(c_i))\cdot g(\mathbf{z}_i) \right ]^{1/k}, \\ \textrm{ where } H(c_i) =-[(g(\mathbf{z}_i) \cdot \log_2 g(\mathbf{z}_i)+(1-g(\mathbf{z}_i)) \cdot \log_2(1-g(\mathbf{z}_i))],\textrm{ and }\\
 g(\mathbf{z}_i) = \frac{1}{n_i} \cdot \sum^{n_i}_{a=1} \left (0.5 + \frac{\cos(\mathbf{z}_i,\mathbf{x}_a)}{2} \right )   
\end{gathered}  
\end{equation}

\subsubsection{Overall Deviation} 
The {overall deviation} ($Dev$) is computed as the overall summed distances between data points and their corresponding cluster center, as shown in Eq.~\ref{deviation}. $Dev$ must be minimized in order to obtain compact clusters~\cite{Handl2005a}.
\begin{equation}\label{deviation}
Dev(\pi) =  \displaystyle\sum_{\mathbf{c}_i \in \pi}\displaystyle\sum_{\mathbf{x}_a \in \mathbf{c}_i}d(\mathbf{x}_a,\mathbf{z}_i)
\end{equation}

\subsubsection{Intra-cluster Variance}   
The {intra-cluster variance} ($Var$) is conceptually similar to that of $Dev$, as shown in Eq.~\ref{var}, and it must be minimized as an objective function~\cite{Garza2018}. \begin{equation}\label{var}
	Var(\pi) =  \frac{1}{n} \cdot \displaystyle\sum_{\mathbf{c}_i \in \pi}\displaystyle\sum_{\mathbf{x}_a \in \mathbf{c}_i}d(\mathbf{x}_a,\mathbf{z}_i)
\end{equation}

\subsubsection{Total Within-Cluster Variance}   
The {total within-cluster variance} ($TWCV$) is also applied to identify sets of compact clusters, as defined in Eq.~\ref{TWCV}, where $f$ is the size of the dimensional feature space, $\mathbf{x}_{ar}$ denotes the $r$th feature value of the $a$th data point, $\mathbf{z}_{ir}$ is the centroid of the $i$th cluster of the $r$th feature. Moreover, $w_{ai}$ (with varying $a$) a generic data point that belongs to the $i$th cluster, where $w_{ai} \in \left[ 0,1\right]$ and $\sum^k_{i=1} w_{ai}=1$. The goal is to minimize $TWCV$ to obtain compact clusters~\cite{Du2005}.
\begin{equation}\label{TWCV}
\begin{gathered}
    TWCV(\pi) = \sum^k_{i=1}\sum^n_{a=1}  {w}_{ai} \sum^f_{r=1}
    (\mathbf{x}_{ar}-\mathbf{z}_{ir})^2, \textrm{where } 
    \mathbf{z}_{ir} = \frac{\sum^n_{a=1} w_{ai}\cdot \mathbf{x}_{ar}}{\sum^n_{a=1}w_{ai}},\\ \textrm{and } w_{ai} \begin{cases}
    {1}, \text{if}\ a^{th} \textrm{object belongs to the } i^{th} cluster\\
    0, \text{otherwise} 
    \end{cases}
\end{gathered}
\end{equation}

\subsubsection{Connectivity Index} 
The {connectivity} index ($Con$)~\cite{Handl2005a}  evaluates the degree to which neighboring data points have been placed in the same cluster. This index is computed according to Eq.~\ref{eq:con}, where $L$ is the parameter that determines the number of nearest neighbors that contribute to the connectivity and $o_{ah}$ is the $h$th nearest neighbor of the object $\mathbf{x}_a$. $Con$ as objectives must be minimized. 
\begin{equation}\label{eq:con}
    Con(\pi) = \displaystyle\sum_{a=1}^{n}\displaystyle\sum_{h=1}^{L}f(\mathbf{x}_a,o_{ah}), \textrm{where} f(\mathbf{x}_a,o_{ah}) = 
    \begin{cases}
    \frac{1}{k},\text{if}\ \nexists\mathbf{c}_i:\mathbf{x}_a, o_{ah}\in\mathbf{c}_i\\
    0, \text{otherwise}
    \end{cases}
\end{equation}

\subsubsection{Data Continuity Degree} 
The {data continuity degree} ($DCD$)  measures the connectedness of the data in terms of the connectivity factor (the total edges sum for each minimum spanning tree) in a similarity graph. In general, it can be computed in two steps.  First, a similarity function is applied in order to generate a similarity graph, the $k_{size}$-Graph. In this graph, a vertex $v_a$ is connected with the vertex $v_b$ if $v_b$ is among the $k$-nearest neighbors of $v_a$. After that, the total minimal spanning tree edges are computed considering all nodes connected within the neighborhood of the current node and internally --- this process is repeated with each connected component due to the graph  not being fully connected. The average arithmetic value of the metric (the connectivity factor divided by the number of clusters) is the result of this objective, which must be maximized in the optimization~\cite{menendez2013multi}.

\subsubsection{Average Between-Group Sum of Squares} 
The {average between-group sum of squares} ($ABGSS$) is computed as the average distance between the centroids of the clusters and the centroid of the data, as defined in Eq.~\ref{abgss}. It must be maximized to obtain well-separated clusters~\cite{Kirkland2011}. 
\begin{equation}\label{abgss}
    ABGSS(\pi) =  \frac{\sum_{i=1}^k n_i \cdot d(\mathbf{z}_i,\overline{\mathbf{z}})}{k}
\end{equation}

\subsubsection{Average Separation Index} 
The {Average Separation} index ($Sep\textsubscript{AL}$) measures the average separation distance between all clusters, as defined in Eq.~\ref{sep_al}~\cite{Ripon2009}. It must be maximized to improve the clustering separation
\begin{equation}\label{sep_al}
    Sep_{AL}(\pi) = \frac{1}{k(k-1)/2} \cdot \sum^k_{i\neq j} d(\mathbf{z}_i, \mathbf{z}_j), 
\end{equation}

\subsubsection{Separation Index} 
The {Separation Index} ($Sep\textsubscript{CL}$) is computed as the sum of the distance between every two tuples (data points) in different clusters, as shown in Eq.~\ref{sep_cl}. It must be maximized to get well-separated clusters~\cite{dutta2012data}.  
\begin{equation}\label{sep_cl}
    Sep_{CL}(\pi) = \sum_{\substack{\mathbf{x}_a \in \mathbf{c}_i,\mathbf{x}_b \in \mathbf{c}_j}, i \neq j} {d(\mathbf{x}_a, \mathbf{x}_b)}
\end{equation}

\subsubsection{Graph-based Separation Index}
The {graph-based separation} index ($Sep_{graph}$) measures the separation between the clusters in terms of  a similarity graph. As in the $DCD$ index, it considers the generation of a $K_{size}$-Graph as the first step in computing this index. The $Sep_{graph}$  is calculated as the arithmetic average value of the edge weights between the different clusters, as defined in Eq.~\ref{sep_graph}, where $\mathbf{c}$ is a cluster, $\mathbf{G}$ is the $K_{size}$-Graph, $\mathbf{v}_a$ is the vertex $a$, and $w_{ab}$ is the edge weight value from node $a$ to node $b$.  $Sep_{graph}$ must be maximized to improve cluster separation~\cite{menendez2013multi}.
\begin{equation}\label{sep_graph}
    Sep_{graph} = \left(\frac{\sum_{\mathbf{v}_a \in \mathbf{G}}\{w_{ab}| \mathbf{v}_a \notin \mathbf{c}\}}{\mathbf{G}-\mathbf{c}}\right) / \mathbf{c}
\end{equation}

\subsubsection{Calinski-Harabasz Index} 
The {Calinski-Harabasz} index ($CH$) considers the degree of dispersion between clusters to measure how similar an object is to its own cluster compared to other clusters. It can take values in [0, $\infty$] with higher values indicating better clustering. $CH$ is computed by the ratio of the sum of between-clusters dispersion and inter-cluster dispersion for all clusters, as shown in Eq.~\ref{CH}~\cite{liu2010understanding}.
\begin{equation}\label{CH}
    CH(\pi) = \frac{\sum\limits_{i=1}^k n_i \cdot d(\mathbf{z}_i,\overline{\mathbf{z}})} {\sum\limits_{i=1}^k \sum\limits_{\mathbf{x}_a\in \mathbf{c}_i} d(\mathbf{x}_a,{\mathbf{z}_i})}\cdot \frac{ (n-k)}{(k-1)}
\end{equation}

\subsubsection{Davies-Bouldin Index} 
The {Davies-Bouldin} index ($DB$) is defined as the average similarity measure of each cluster with its most similar cluster based on the ratio of the sum of within-cluster scatter to between-cluster separation ($R_i$). $DB$ is defined in Eq. \ref{DB}, where $S_i$ is the scatter within the $i$th cluster~\cite{liu2010understanding}.
The minimum value of this $DB$ is zero, with lower values indicating better clustering.
\begin{equation}\label{DB}
\begin{gathered}
DB(\pi) = \frac{1}{k} \cdot  \sum_{i=1}^k \max\limits_{j \neq i}\left \{ \frac{S_i+S_j}{d(\mathbf{z}_i, \mathbf{z}_j)} \right \}, \textrm{ where }  S_i=\frac{1}{|n_i|} \cdot \sum _{\mathbf{x}_a  \in \mathbf{c}_i} d(\mathbf{x}_a,\mathbf{z}_i), 
\end{gathered}
\end{equation}

\subsubsection{Dunn index} 
The {$Dunn$} index is computed as the ratio between the minimum inter-cluster distance ($\delta (\mathbf{c}_i,\mathbf{c}_j)$) to the maximum cluster diameter ($\max _{j\leq i \leq k} {\Delta (\mathbf{c}_i)}$), as defined in Eq. (\ref{Dunn}). It is considered that compact and well-separated clusters have a small diameter and a large distance between them. The Dunn index can take values between zero and infinity, and it must be maximized to obtain a well-separated and compact cluster~\cite{liu2010understanding}. 
\begin{equation}\label{Dunn}
\begin{gathered}
    Dunn(\pi) =  \min\limits_{1\leq i\leq k}\left \{\min\limits_{\substack{1\leq j\leq k, \\ j\neq i}}\left   \{ \frac{\delta (\mathbf{c}_i,\mathbf{c}_j)}{\max\limits_{j\leq i \leq k} {\Delta (\mathbf{c}_i)}}   \right \}  \right \},  \\\textrm{ where } \delta (\mathbf{c}_i, \mathbf{c}_j) =   \min\limits_{\substack{\mathbf{x}_a \in \mathbf{c}_i,\\ \mathbf{x}_b \in \mathbf{c}_j}}\left \{ d(\mathbf{x}_a,\mathbf{x}_b) \right \},  \textrm{ and }  \Delta (\mathbf{c}_i)=\max\limits_{\mathbf{x}_a,   \mathbf{x}_b \in \mathbf{c}_i} \{d(\mathbf{x}_a,\mathbf{x}_b)\}
\end{gathered}
\end{equation}

\subsubsection{Modularity index} 
The {modularity} index ($Mod$) is computed as the total difference between the sum of distances of the objects in the same cluster $\mathbf{c}_i$ (that indicates how closely data similar with others in the same cluster) and the sum of distances considering the objects in the dataset $\mathbf{X}$ (that determines how closely data similar with others in different clusters), as defined in Eq. \ref{eq:mod}~\cite{Liu2018}.  
  \begin{equation}\label{eq:mod}
Mod(\pi) =  \sum^k_{i=1}\left[\left(\frac{\sum\limits_{\substack{ \mathbf{x}_a,\mathbf{x}_b \in \mathbf{c}_i}}d(\mathbf{x}_a,\mathbf{x}_b)}{\sum\limits_{\mathbf{x}_a,\mathbf{x}_b \in \mathbf{X}}d(\mathbf{x}_a,\mathbf{x}_b)} \right)
- \left(\frac{ \sum\limits_{\substack{\mathbf{x}_a \in \mathbf{c}_i,\mathbf{x}_b \in \mathbf{X} }}d(\mathbf{x}_a,\mathbf{x}_b)}{\sum\limits_{\mathbf{x}_a,\mathbf{x}_b \in \mathbf{X}}d(\mathbf{x}_a,\mathbf{x}_b)} \right)^2 \right]
\end{equation}

\subsubsection{Silhouette Index} 
The {Silhouette} index ($Sil$) measures how much each point in the data is similar to its own cluster compared to other clusters, based on the relation of the mean similarity of the objects within a cluster and the mean distance to the objects in the other clusters.  $Sil$ is defined in Eq. \ref{sil}, in which $ad_a$ refers to the mean distance between a sample $\mathbf{x}_a$ and all other points in the same cluster. Moreover, $bd_a$ is the mean distance between a sample $\mathbf{x}_a$ and the nearest cluster that  $\mathbf{x}_a$ is not a part of. Thus, $Sil$ produces values between $-1$ and $1$. A higher value corresponds to a better clustering result~\cite{Mukhopadhyay2007}.  
\begin{equation}\label{sil}
Sil(\pi) =   \frac{1}{n}\cdot \sum_{a=1}^n S(\mathbf{x}_a), \textrm{where } S(\mathbf{x}_a) =   \frac{bd_a - ad_a}{max\left \{ ad_a,bd_a \right \}},     
\end{equation}

\subsubsection{Pakhira-Bandyopadhyay-Maulik Index}
 The Pakhira-Bandyopadhyay-Maulik index ($PBM$) is defined in Eq. \ref{pbm}, where $E$ measures the total within-cluster scatter, $E_0$ is the total scatter considering all the samples belonging to one single cluster, and $D$ is the maximum distance between cluster centers. Furthermore, $\mu_{ai}$ denotes the membership degree of the objects in a cluster, which can take values between 0 and 1. In our experiments, we considered a hard clustering, in which each object either belongs to a cluster completely ($\mu_{ai}=1$) or not ($\mu_{ai}=0$). The $PBM$ must be maximized as objective function.
\begin{equation}\label{pbm}
\begin{gathered}
PBM=\frac{1}{k} \cdot\frac{E_0}{E_k} \cdot D_k 
\textrm{ where } E_0 = \sum_{a=1}^n d(\mathbf{x}_a,\mathbf{\overline{z}}), E_k = \sum_{i=1}^k E_i, \\E_i = \sum_{a=1}^n\sum_{i=1}^k \mu_{ai}\cdot d(\mathbf{x}_a,\mathbf{c}_i)^2,\textrm{and }  D_k = \max\limits_{i,j=1, i\neq j}^k d(\mathbf{z}_i,\mathbf{z}_j)
\end{gathered}
\end{equation}

\subsubsection{Xeni-Beny} 
The {Xeni-Beny} index ($XB$) is defined as a function of the ratio of the total fuzzy cluster variance to the minimum separation of the clusters, as defined in Eq. \ref{XB}, where $\mu_{ai}$ is the membership degree of the $a$th data point to the $i$th cluster, and $m$ is the fuzzy exponent. In the same way as in the $PMB$ we considered a hard clustering to define the $\mu_{ai}$~\cite{liu2010understanding}. $XB$ must be minimized to obtain well-separated and compact clusters. 
\begin{equation}\label{XB}
   XB(\pi)    =\frac{\sum\limits_{i=1}^k\sum\limits_{a=1}^n \mu_{ai}^m \cdot d(\mathbf{z}_i, \mathbf{x}_a)} {n\cdot(\min_{i \neq j} \left\{ d(\mathbf{z}_i, \mathbf{z}_j)\right\} )}, 
\end{equation}

\subsection{Clustering Algorithms applied in the EMOCs Initialization}
Here, we present five clustering algorithms: $k$-means, average linkage, single linkage, shared nearest neighbor-based clustering, and minimum spanning tree clustering.  
These clustering algorithms provide different strategies that allow us to evaluate how they can affect the optimization.

\subsubsection {$k$-means}
$k$-means (KM)~\cite{macqueen1967some} is a partitional clustering algorithm applied to detect compact clusters. Its objective is to minimize the distance between the centroid and their respective instances. The $k$-means starts by choosing a $k$ set of centroids randomly (or based on prior knowledge and associating each object with the nearest centroid), where $k$ is a user-given parameter. After that, the centroids are recomputed based on the current cluster data, followed by a new association of each instance with the nearest centroid; this operation is successively repeated until there is no change in the groups or the stopping criterion is met.

\subsubsection {Average linkage and Single linkage}
Average linkage (AL) and Single linkage (SL) are hierarchical algorithms applied to detect nested or hierarchical data structures.  Each instance starts out standing as an individual cluster in both algorithms, and a sequence of merge operations is executed until it reaches a single cluster with all the instances. The core difference between AL and SL is the distance measure used to compute proximity between the pairs of clusters. This measure is used to define the closest pair of sub-sets that are merged. SL uses the minimal distance between two instances of a cluster pair, and AL applies the average distance of all observations of the clusters pairs \cite{xu2005survey}.

\subsubsection {Shared Nearest Neighbor-based clustering}
Shared nearest neighbor-based clustering (SNN)~\cite{ertoz2002new} is a density-based algorithm. SNN can detect clusters of different sizes, shapes, and densities. In general, the main idea behind this algorithm is to use the concept of similarity based on the shared nearest neighbor. The objects are assigned to a cluster that shares a large number of their nearest neighbors (the density based on the neighborhood). 
 This algorithm begins with the computation of the similarity matrix, which is sparsified by retaining only the k-Nearest Neighbors (KNN).  In the following, the shared nearest neighbor graph is constructed, in which links are created between pairs of objects that have each other in their KNN lists. Then, SNN computes the number of shared neighbors between vertices, considering the links coming from each point in the graph, providing the density factor.
This factor is used to identify the noise or core points based on the user-defined thresholds. Then, noisy points are discarded, and the clusters are formed by the core points and the border points (non-noise non-core points) considering all the connected components. 

\subsubsection {Minimum Spanning Tree clustering}
The minimum spanning tree (MST)-clustering is a graph-based algorithm that can identify clusters of arbitrary shapes. Among a variety of versions of this algorithm, we consider here the MST-clustering described in \cite{Handl2005c}. 
This algorithm uses the concept of \textit{degree of interestingness} (DI) and the properties of the MST to find the clusters. The DI defines the neighborhood  relationship between the nodes in the MTS, where a link between two nodes is considered interesting if neither of them is a part of the other node’s set of nearest neighbors. Thus, the clusters are generated by removing interesting links in the MST that split it into sub-graphs in which the connected elements represent a cluster. 

\subsection{Evolutionary Multi-objective Clustering Approaches}\label{sec:emoc}
In the literature, there are a variety of works that apply multi-objective optimization to clustering. For example,~\cite{gong2013complex} applied a multi-objective Particle Swarm approach to network clustering, \cite{saini2019automatic} used the multi-objective differential evolution approach to document clustering,~\cite{shang2019multi} applied the multi-objective artificial immune algorithm for fuzzy clustering,  among other works that used different meta-heuristics (as Artificial Immune system-inspired, Differential Evolution-based, Simulated Annealing-based, Particle swarm-based).  In this paper, we adopted the general classification of the metaheuristics presented by~\cite{siarry2016simulated} to select articles with evolutionary optimization (including memetic and hybrid approaches).  

As above-mentioned, in this study, the objective functions and initialization strategies analyzed were obtained from Evolutionary Multi-Objective Clustering approaches (EMOCs) found in the literature. In the following, we briefly present some aspects of the selected studies. It is important to note that we are focused on analyzing the objective functions and evaluating them regarding the impact of the initial populations in the optimization. Thus, we will not be concerned about the other aspects of the EMOCs (representation, evolutionary operators, and selection method).   

We summarize the information regarding the EMOCs used in this paper in Table~\ref{tab:emocs}. These EMOCs use different initializations. In~\cite{Du2005,Ripon2006_IJCNN,Mukhopadhyay2007,Ripon2009,Mukhopadhyay2009,Kirkland2011,menendez2013multi,menendez2014co}, they use a random initialization. 
As pointed out in \cite{hruschka2009survey}, the random initialization generally provides unfavorable partitions since the clusters are likely to be mixed up to a high degree. However, this strategy is very popular because of its simplicity and effectiveness in testing the algorithms against hard evaluation scenarios.   
On the other hand, the other approaches~\cite{Handl2005b,Handl2005c,Faceli2006,Handl2007,Matake2007,Tsai2012,Garza2018,Zhu2018,Liu2018,zhu2020evolutionary} consider high-quality partitions in the initialization generated by clustering algorithms. 
In this article, we evaluate this second strategy, considering the initialization of established EMOCs: MOCLE (Multi-Objective Clustering Ensemble)~\cite{Faceli2006} and  MOCK (Multi-Objective Clustering with automatic k-determination)~\cite{Handl2005c}.
As for MOCLE~\cite{Faceli2006}, it is not linked to a specific initialization strategy, in which it is possible to use different clustering algorithms to generate the initial population; thus, in this paper, we applied the same initialization used in~\cite{Faceli2006}, which uses four clustering algorithms: KM, AL, SL, and SNN. MOCK~\cite{Handl2005c} and some derived works presented in~\cite{Matake2007,Tsai2012} considered the initialization based on MST-clustering and KM. MOCK-medoid~\cite{Handl2005b} uses only the KM in the initialization. Most recent works derived from MOCK, such as those in~\cite{Garza2018,Zhu2018,zhu2020evolutionary}, rely solely on MST-clustering to generate the initial population. 

\begin{table}[htb]
 \caption{Multi-objective clustering approaches}	
 \centering
\scalebox{0.7}{
	\begin{tabular}{|l|l|l|l|}	\hline
	\textbf{Year} & \textbf{EMOC} & \textbf{Initialization} & \textbf{Objective Functions}\\ \hline
	2005 & MOCK \cite{Handl2005c}  & MST-clustering and KM&$Con$ and $Dev$ \\\hline
	2005 & MOCK-medoid \cite{Handl2005b}  & KM&$Var$ and $Dev$ \\\hline
	2005 & MOGA-LL \cite{Du2005}  & Random & $TWCV$ and $k$  \\\hline
	2006 & VRJGGA \cite{Ripon2006_IJCNN} & Random  & $Ent$ and $Sep\textsubscript{AL}$ \\\hline
	2006 & MOCLE \cite{Faceli2006} &  KM, AL, SL, SNN & $Con$ and $Dev$ \\\hline
	2007 & MOGA-medoid~\cite{Mukhopadhyay2007}& Random & $Dev$ and $Sil$  \\\hline
	2007 & MOCK-scalable \cite{Matake2007}& MST-clustering and KM  &  $Con$ and $Dev$ \\\hline
	2009 & EMCOC \cite{Ripon2009} & Random & $Ent$ and $Sep\textsubscript{AL}$\\\hline
	2009 & MOVGA \cite{Mukhopadhyay2009} & Random &$PBM$ and $XB$\\\hline
	2011 & MOCA \cite{Kirkland2011} & Random  & $AWGSS$, $ABGSS$ and $Con$ \\	\hline
	2012 & MIE-MOCK \cite{Tsai2012}  & MST-clustering and KM & $Con$ and $Dev$\\\hline
	2013 & MOGGC \cite{menendez2013multi} & Random & $sep_{graph}$ and $DCD$ \\\hline
	2014 & CEMOG \cite{menendez2014co} & Random  & $sep_{graph}$ and $DCD$ \\ \hline
	2017 & $\Delta$-MOCK \cite{Garza2018} & MST-clustering  & $Con$ and $Var$ \\\hline
	2018 & $\Delta$-EMaOC \cite{Zhu2018}& MST-clustering& $Con$, $Var$, $Dunn$, $DB$ and $CH$ \\\hline
	2018 & MOECDM \cite{Liu2018} & Random and NCUT & $Sep_{CL1}$ and $Sep_{CL2}$  \\\hline
	2018 & MOEACDM \cite{Liu2018} & NCUT  & $Mod_1$ and $Mod_2$  \\\hline	
	2020 & MOAC-L\cite{zhu2020evolutionary} & MST-clustering & $Con$ and $Var$\\\hline
	
\end{tabular}}
\label{tab:emocs}
\end{table}

Regarding the objective functions, these EMOCs use different combinations of criteria. A popular pair of objective functions is $Dev$ (compactness criterion) and $Con$ (connectedness criterion) used in MOCK~\cite{Handl2005c} and other derived works presented in \cite{Matake2007,Tsai2012}. Recent works derived from MOCK~\cite{Garza2018,Zhu2018,zhu2020evolutionary} use $Var$ instead of $Dev$ as compactness criterion. In~\cite{Zhu2018}, the authors included three objective functions:  $Dunn$, $DB$, and $CH$, applied along with $Var$ and $Con$. On the other hand, $TWCV$ and the number of clusters $k$ are optimized in the approach presented by~\cite{Du2005}.
A particular case occurred in MOCK-medoid~\cite{Handl2005b}, in which $Dev$ and $Var$, two very similar compactness criteria, are applied in order to develop a multi-objective clustering around medoids, where the computation of cluster medoids does not require the presence of feature vectors but can be done for dissimilarity data. In a similar way, $Dev$ and $Sil$ are the objectives employed by~\cite{Mukhopadhyay2007}.

In~\cite{Kirkland2011} $Con$ is also applied. However, the authors consider the combination with the Average Within Group Sum of Squares ($AWGSS$) and $ABGSS$. This paper does not conduct experiments with $AWGSS$ because it is very similar to $Var$ and $Dev$ in its formulation to detect compact clusters --- in fact $AWGSS$ measures the average distance between each object in the cluster and its centroid. 
In \cite{menendez2013multi,menendez2014co}, a connectedness criterion, i.e. $CDC$, is associated with $sep_{graph}$, a separation criterion. 
Other works, such as \cite{Ripon2006_IJCNN,Ripon2009}, consider a separation criterion, $Sep_{Al}$, and a compactness criterion, $Ent$, as objective functions. Moreover, \cite{Liu2018} presents the use of one criterion (separation or compactness) considering two different distance functions. For example, they use $Sep_{CL}$ as an objective function, in which one objective considers the Euclidean distance and the other one the path distance. 

It is important to note that besides the EMOCs presented in Table~\ref{tab:emocs}, there are other EMOCs in the literature, such as~\cite{dutta2012data,shang2016multiobjective,dutta2019automatic}, among others. These works use a random initialization. Besides that, ~\cite{dutta2012data,dutta2019automatic} use a pair of objective functions considering compactness and separation criteria that are very similar in formulation to other selected clustering criteria. In contrast,~\cite{shang2016multiobjective} presented objective functions applied to a specific application (community detection). These two cases are not considered in our experiments.

\section{Experimental Design}\label{sec:experiment}

\subsection{Goals of the experiments}\label{sec:goalexp}
Aiming to support the choice of the objective functions to design the EMOCs, in this paper we propose an analysis of the admissibility of the objective functions. Therefore, the specific goals of our experiments are to answer the following research questions:(i) ``Are there existing objective functions that can lead the EMOCs to find the optimal value (true partition)?'', (ii) ``Are there specific features or optimizing scenarios that should be considered in the choice and combination of the objective functions to obtain better clustering results?''. 

\subsection{Experimental setup}\label{sec:goalexpsep}
In order to answer the first research question, we evaluated the admissibility of each objective function presented in Section~\ref{sec:obj}, considering the base partitions obtained from different initialization strategies. 
Therefore, we used the initialization algorithm of the MOCK and MOCLE, both established and popular approaches, that consider different clustering criteria.    
Thus, we generated five initial populations using the clustering algorithm: KM, AL, SL, SNN, and MST-clustering. The general setting applied in the KM, AL, SL, and SNN is the same as reported in \cite{Faceli2006}. Regarding the MST-clustering, we employed the general setting presented in \cite{Handl2007}. Furthermore, we adjusted such algorithms to produce partitions containing clusters in the range $\{2,2k^{*}\}$, where $k^*$ is the number of clusters in the true partition representing the dataset. This setting is commonly used in MOCK/$\Delta$-MOCK’s to define the number of clusters in the partitions of the initial population.
In this initial experiment, in particular, we analyze the admissibility by comparing the individuals of the initial populations with the optimal value (true partition) of each dataset. The results of this experiment provide us with information about which objective function and initialization strategy could improve the optimization. 

Regarding the second research question, we evaluated different combinations of objective functions and analyzed which conditions could lead the EMOC to provide better results. In particular, we analyze the clustering performance of some promising objective functions found in the first experiment.  Experiments were carried out in the new MOCK version, $\Delta$-MOCK~\cite{Garza2018}.  We select this EMOC, among others, because it is a recently established approach in which the present features (as the use of MST-clustering in the initialization) contribute to the evaluation in terms of the search direction, demonstrating how admissibility supports the choice of objective functions.
Regarding this algorithm setting, we employed the one reported in~\cite{Garza2018}, considering the $\Delta$-locus scheme with $\delta$ settled heuristic~$\sim5/\sqrt{n}$, where $n$ is the number of objects in the dataset. 

\subsection{Datasets} 
As previously stated, our analysis takes into account the use of the true partition. Thus, we selected 24 artificial datasets, in which we can analyze the relationship between their data structures or cluster shapes and the optimization of the objective functions. Table~\ref{table:characteristics} summarizes the main characteristics of these datasets, in which $n$ is the number of objects, $d$ refers to the number of attributes (dimensions), and $k$* is the number of clusters in the true partition.

\begin{table}[ht]
\centering
	\begin{subtable}[h]{0.45\textwidth}
	\centering
	\scalebox{0.7}{
	\begin{tabular}{l|l|r|r|r}
		\hline
		  \textbf{G} & \textbf{Dataset} & \textbf{$n$} & \textbf{$d$} &  $k^{*}$ \\		\hline
			  \multirow{12}{*}{G1}&\texttt{R15} & 600 & 2 & 15 \\
        &\texttt{D31} & 3.100 & 2 & 31 \\
				&\texttt{Engytime} & 4.096 & 2 & 2 \\
				&\texttt{Sizes5} & 1.000 & 2 & 4 \\
				&\texttt{Square1} & 1.000 & 2 & 4 \\
				&\texttt{Square4} & 1.000 & 2 & 4 \\
				&\texttt{Twenty} & 1.000 & 2 & 20  \\ 
				&\texttt{Fourty} & 1.000 & 2 & 40 \\
				&\texttt{Sph\_5\_2} & 250 & 2 & 5 \\
				&\texttt{Sph\_6\_2} & 300 & 2 & 6 \\
				&\texttt{Sph\_9\_2} & 900 & 2 & 9 \\
				&\texttt{Sph\_10\_2} & 500 & 2 & 10 \\	\hline
		\end{tabular}}
   \end{subtable}
    \hspace{3em}%
    \begin{subtable}[h]{0.45\textwidth}
			\scalebox{0.7}{
			\begin{tabular}{l|l|r|r|r}\hline
				\textbf{G} &\textbf{Dataset} & \textbf{$n$} & \textbf{$d$} &$k^{*}$ \\ \hline
					\multirow{3}{*}{G2}&\texttt{ds2c2sc13\_S1} & 588 & 2 & 2 \\
					&\texttt{ds2c2sc13\_S2} & 588 & 2 & 5 \\
					&\texttt{ds2c2sc13\_S3} & 588 & 2 & 13 \\ \hline
					\multirow{3}{*}{G3}&\texttt{Long1} & 1.000 & 2 & 2 \\
					&\texttt{Pat2} & 417 & 2 & 2 \\
					&\texttt{Spiral} & 1.000 & 2 & 2 \\ \hline
					\multirow{6}{*}{G4} &\texttt{3MC} & 400 & 2 & 3 \\
					&\texttt{DS-850} & 850 & 2 & 5 \\
					&\texttt{Aggregation} & 788 & 2 & 7 \\
					&\texttt{Complex9} & 3.030 & 2 & 9 \\
					&\texttt{Pat1} & 557 & 2 & 3 \\
					&\texttt{Spiralsquare} & 2.000 & 2 & 6 \\ \hline  
		\end{tabular}}
	\end{subtable}
	\caption{Dataset characteristics} 
	\label{table:characteristics}
\end{table}

\begin{figure}[!htb]
	\centering
	\subfloat[\texttt{R15}]{\includegraphics[width=0.315\textwidth]{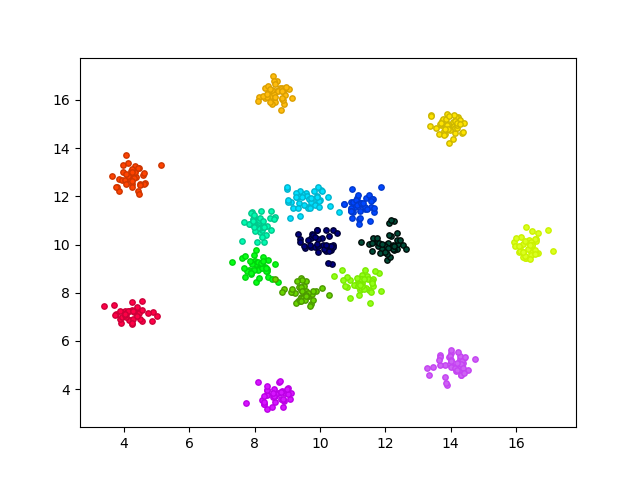}\label{R15}}
		\hfill
  \subfloat[\texttt{D31}]{\includegraphics[width=0.315\textwidth]{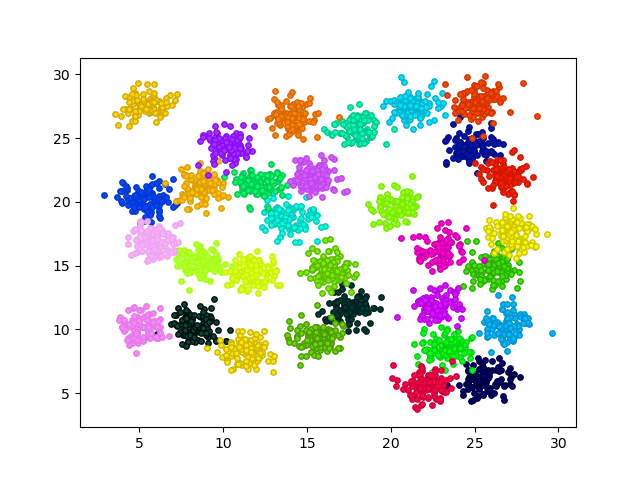}\label{D31}}
	\hfill
	\subfloat[\texttt{engytime}]{\includegraphics[width=0.315\textwidth]{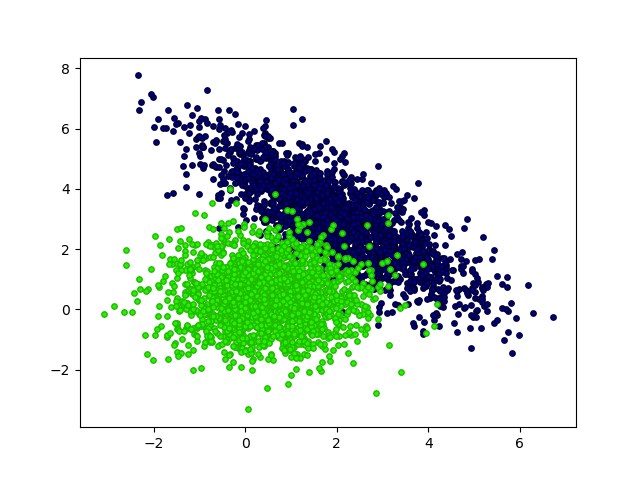}\label{energytime}}
	\hfill
  \subfloat[\texttt{Sizes5}]{\includegraphics[width=0.315\textwidth]{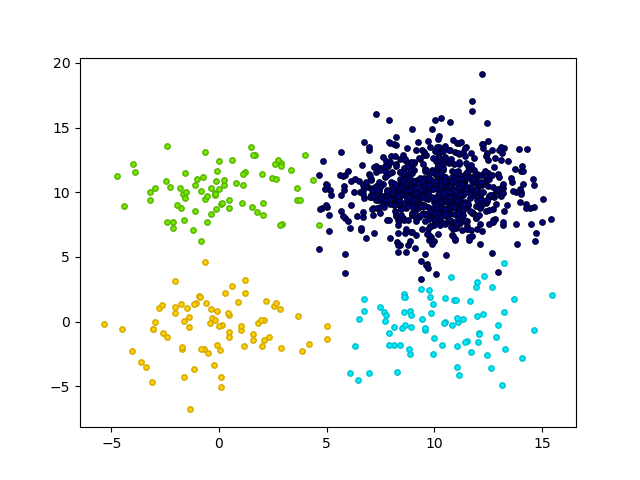}\label{Sizes5}}
  \hfill
	\subfloat[\texttt{Square1} ]{\includegraphics[width=0.315\textwidth]{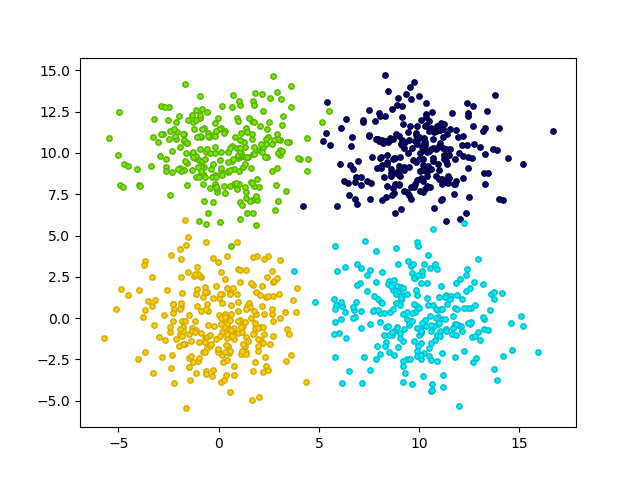}\label{Square1}}
	\hfill
  \subfloat[\texttt{Square4}]{\includegraphics[width=0.315\textwidth]{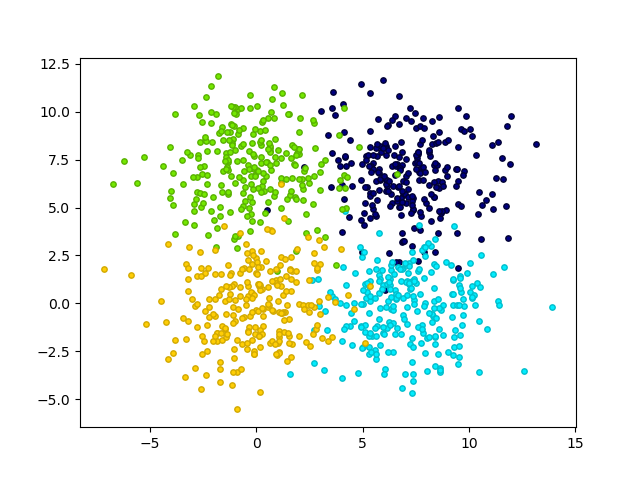}\label{Square4}}
	  \hfill
	\subfloat[\texttt{Fourty} ]{\includegraphics[width=0.315\textwidth]{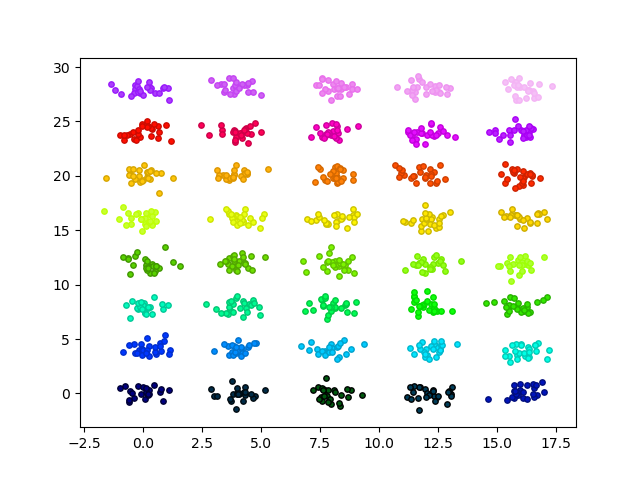}\label{Fourty}}
	\hfill
  \subfloat[\texttt{Twenty}]{\includegraphics[width=0.315\textwidth]{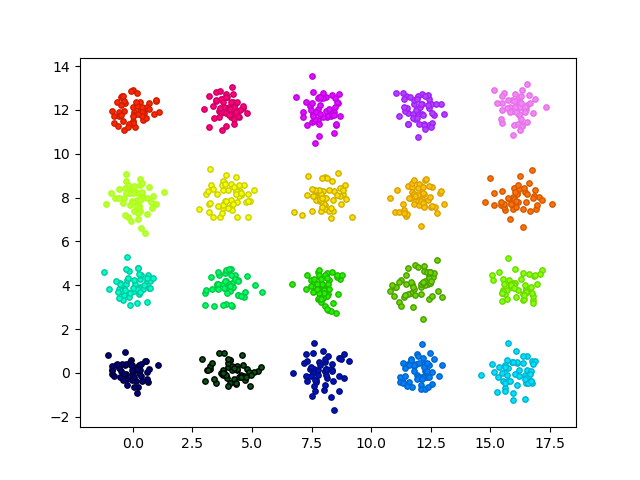}\label{Twenty}}
	\hfill
	\subfloat[\texttt{Sph\_5\_2}]{\includegraphics[width=0.315\textwidth]{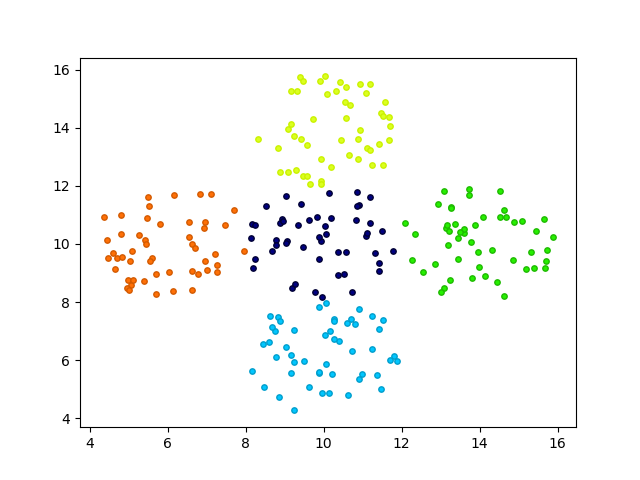}\label{Sph52}}
	\hfill
	\subfloat[\texttt{Sph\_6\_2}]{\includegraphics[width=0.315\textwidth]{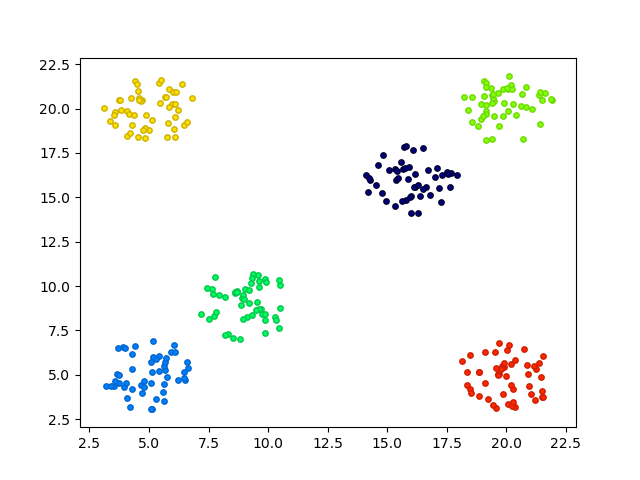}\label{Sph62}}
	\hfill
  \subfloat[\texttt{Sph\_9\_2}]{\includegraphics[width=0.315\textwidth]{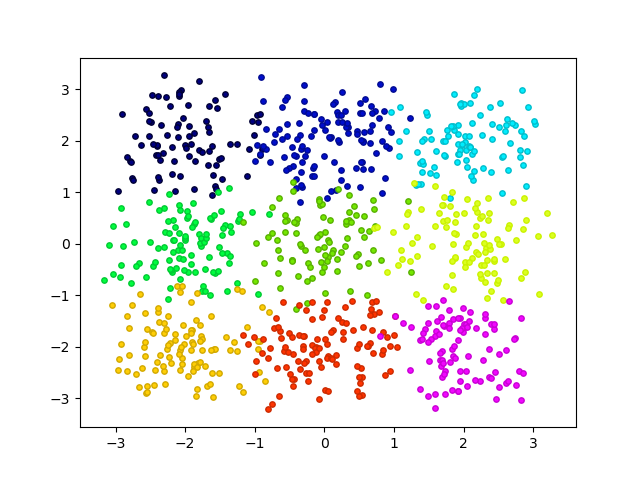}\label{Sph92}}
	\hfill
  \subfloat[\texttt{Sph\_10\_2}]{\includegraphics[width=0.315\textwidth]{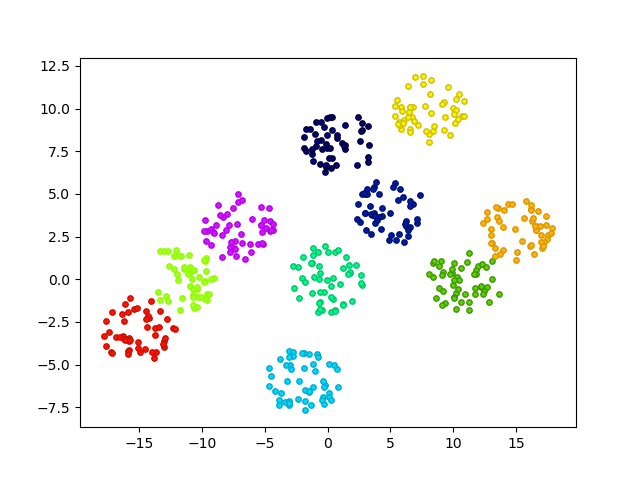}\label{Sph102}}
    \caption{{Datasets with gaussian-like and hyper-spherical shaped clusters}}
	\label{fig:datasets1}
\end{figure}

We divided these datasets into four groups (column $\mathbf{G}$ in Table \ref{table:characteristics}), considering similar data structures evaluated in our analysis. In the first group (G1), Fig.~\ref{fig:datasets1}, we have 8 datasets with gaussian-like clusters and 4 datasets with hyper-spherical shaped clusters.  \texttt{R15}, \texttt{D31}, \texttt{Engytime},  \texttt{Sizes5}, \texttt{Square1}, \texttt{Square4}, \texttt{Twenty}, and \texttt{Fourty} have gaussian-like clusters. 
 \texttt{R15} consists of 15 identical-sized clusters with some overlapping points. \texttt{D31} has 31 clusters that are slightly overlapping and distributed randomly. \texttt{Engytime} has two highly overlapping clusters with different variances. 
\texttt{Size5} has five clusters of varying sizes and the same inter-cluster distance over all clusters. 
 \texttt{Square1} and \texttt{Square4} consist of four clusters of equal size and spread that vary in the degree of overlap and the relative size of clusters. \texttt{Fourty} and \texttt{Twenty} consist of well-separated small clusters distributed in 40 and 20 clusters, respectively.  
\texttt{Sph\_5\_2}, \texttt{Sph\_6\_2}, \texttt{Sph\_9\_2}, \texttt{Sph\_10\_2} have hyper-spherical shaped clusters with different proximity between the clusters.  Algorithms based on cluster compactness, such as KM, can detect well-separated hyper-spherical shaped clusters; they can also detect gaussian-like clusters when they contain globular (no oblong) and well-separated data structures. 

\begin{figure}[!htb]
	\centering
  \subfloat[\texttt{S1}]{\includegraphics[width=0.315\textwidth]{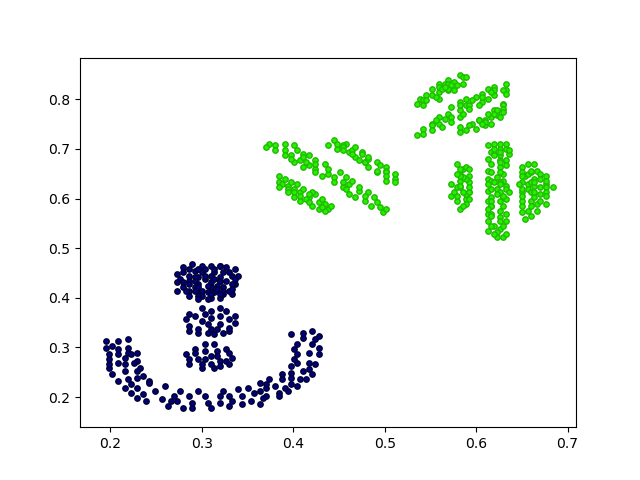}\label{ds2c2sc13V1}}
	\hfill
  \subfloat[\texttt{S2}]{\includegraphics[width=0.315\textwidth]{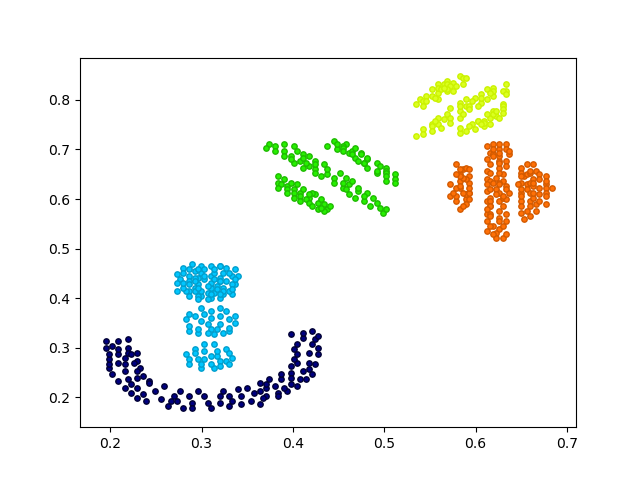}\label{ds2c2sc13V2}}
	\hfill
  \subfloat[\texttt{S3}]{\includegraphics[width=0.315\textwidth]{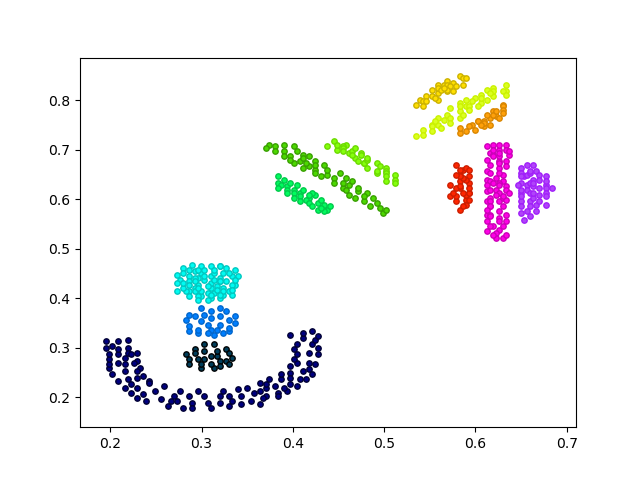}\label{ds2c2sc13V3}}
    \caption{{ds2c2sc13 data structures}}
	\label{fig:datasets2}
\end{figure}

In the second group (G2), Fig.~\ref{fig:datasets2}, we have the \texttt{ds2c2sc13} dataset, which contains three different structures: \texttt{S1}, \texttt{S2}, and \texttt{S3}. These structures represent three levels of structures in a nested dataset. In this example, {\tt S1} represents two well-separated clusters, which can be found by techniques based on optimizing connectedness or compactness; in contrast, {\tt S2} and {\tt S3} combine distinct types of clusters that could be hard to find with techniques based only on connectedness or compactness. Hierarchical clustering algorithms, such as SL and AL, are usually applied to detect nested structures.   

\begin{figure}[!htb]
  \subfloat[\texttt{Long1}]{\includegraphics[width=0.315\textwidth]{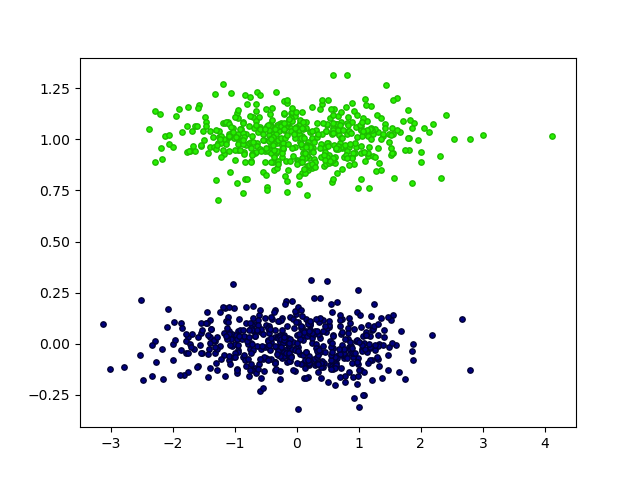}\label{Long1}}
	\hfill
  \subfloat[\texttt{Pat2}]{\includegraphics[width=0.315\textwidth]{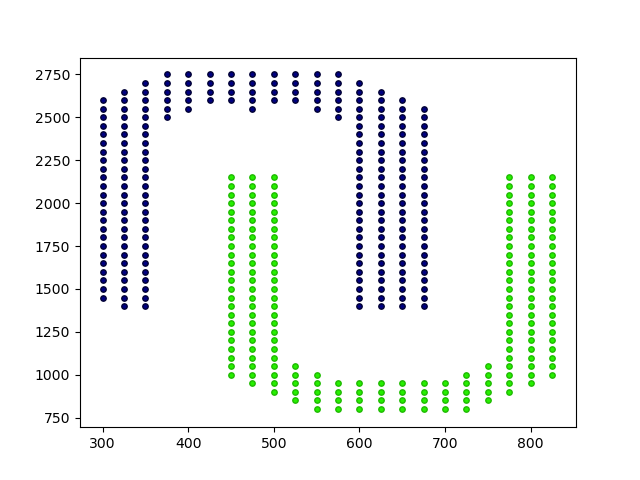}\label{Pat2}}
	\hfill
  \subfloat[\texttt{spiral}]{\includegraphics[width=0.315\textwidth]{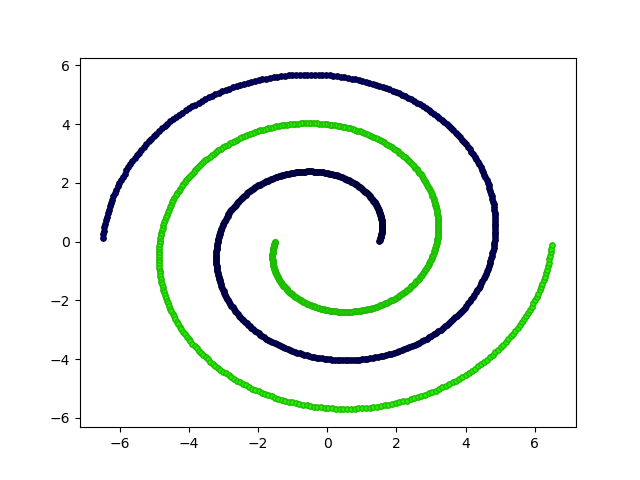}\label{spiral}}
    \caption{{Datasets with elongated cluster shapes}}
	\label{fig:datasets3}
\end{figure}

In the third group (G3), Fig~\ref{fig:datasets3}, we have datasets that contain well-separated and elongated cluster shapes that are hard to identify for algorithms based on cluster compactness: \texttt{Long1}, \texttt{Spiral}, and \texttt{Pat2}. 

\begin{figure}[!htb]
	\centering
  \subfloat[\texttt{3MC}]{\includegraphics[width=0.315\textwidth]{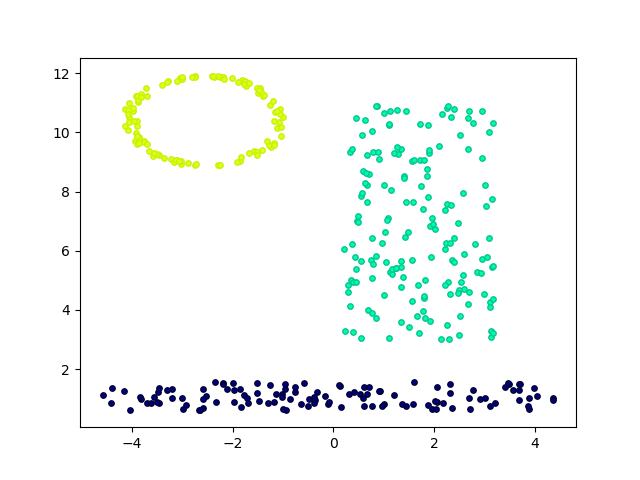}\label{3MC}}
	\hfill
  \subfloat[\texttt{DS-850}]{\includegraphics[width=0.315\textwidth]{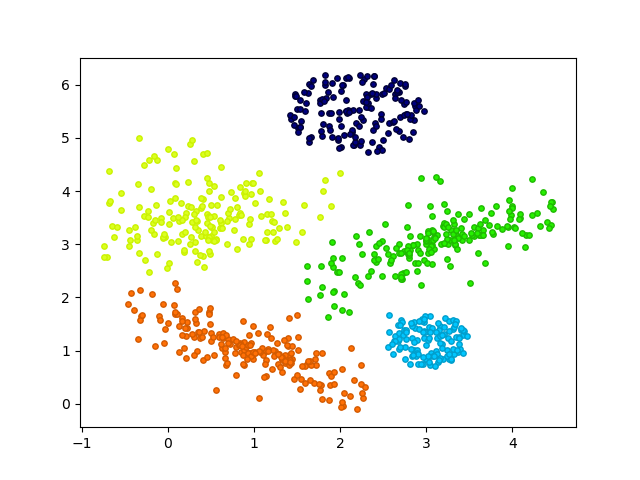}\label{DS-850}}
	\hfill
  \subfloat[\texttt{Aggregation}]{\includegraphics[width=0.315\textwidth]{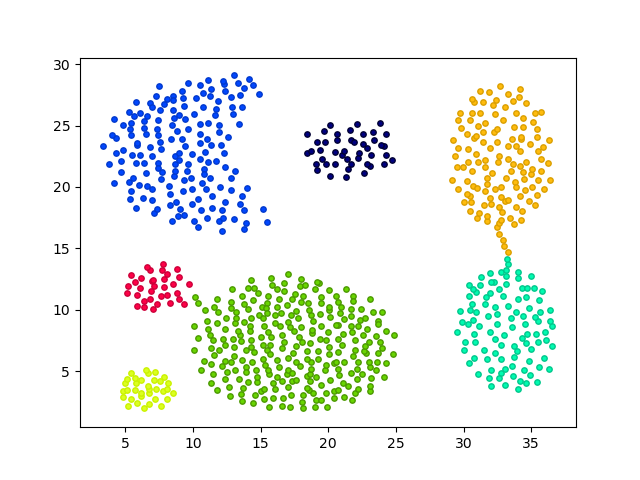}\label{Aggregation}}
	\hfill
  \subfloat[\texttt{complex9}]{\includegraphics[width=0.315\textwidth]{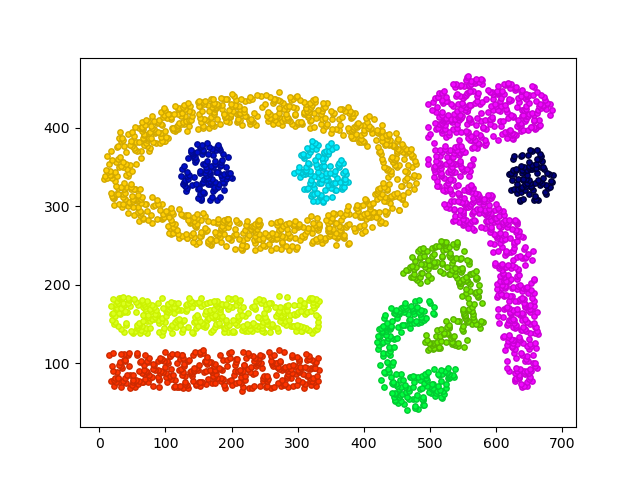}\label{complex9}}
	\hfill
  \subfloat[\texttt{Pat1}]{\includegraphics[width=0.315\textwidth]{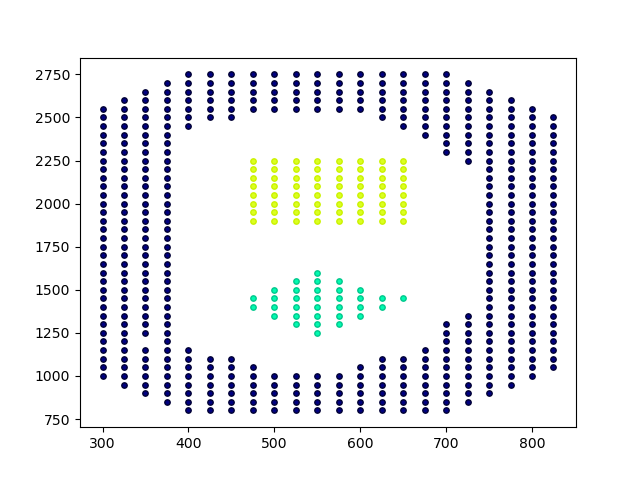}\label{Pat1}}
	\hfill
  \subfloat[\texttt{spiralsquare}]{\includegraphics[width=0.315\textwidth]{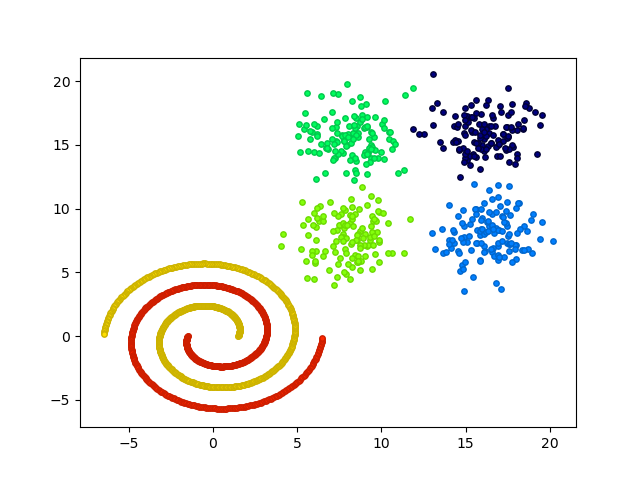}\label{spiralsquare}}
    \caption{{Datasets with distinct types of clusters}}
	\label{fig:datasets4}
\end{figure}

In the last group (G4), Fig.~\ref{fig:datasets4}, we have shaped datasets that combine different types of clusters: \texttt{3MC}, \texttt{DS-850}, \texttt{Aggregation}, \texttt{Complex9}, \texttt{Pat1}, \texttt{Spiralsquare}. \texttt{Aggregation} contains 6 clusters with a uniform and compact distribution, and they also have different sizes, and two clusters are linked by a line of points. \texttt{3MC} contains symmetrical shaped clusters (e.g., ring-shape, ellipsoidal clusters, etc.). \texttt{Pat1} and \texttt{Complex9} present clusters surrounding other ones, among other data structures. \texttt{Spiralsquare} combines spirals and square shapes  into clusters. 

It is important to observe that the use of artificial datasets that provide the true partition and also have well-known data structures makes it possible to analyze in detail the conditions that can affect the optimization in our study. We consider that our study can be a reference base for future analysis considering real-life datasets.

\subsection{Performance assessment}
In terms of optimization, we evaluated the objective function’s admissibility and analyzed the solutions regarding the dominance of the true partition. The general concepts of both these items (admissibility and dominance) are presented in Section~\ref{section:background}.
As main indicator of clustering performance, we used the adjusted Rand index (ARI)~\cite{hubert1985comparing}. The ARI measures the similarity between two partitions, in which results close to 0 express no agreement between the partitions, and results close to 1 indicate high similarity between the partitions.  
ARI is defined in Eq.~\ref{eq:RC}, where $k^a$ and $k^b$ are the numbers of clusters of partitions $\pi^a$ and $\pi^b$, $c^a_i$ corresponds to the i-th cluster from partition $\pi^a$,  $c^b_j$ is the j-th cluster from $\pi^b$ and $n$ is the number of objects in the dataset. 
\begin{equation}
ARI = \frac{\sum_{i=1}^{k^a} \sum ^{k^b} _{j=1} \binom{|c^a_i\cap c^b_j|}{2} - \begin{bmatrix}
 \sum ^{k^a}_{i=1} \binom{|c^a_i|}{2} \sum ^{k^b} _{j=1} \binom{|c^b_j|}{2} \end{bmatrix}/\binom{n}{2}}
{\begin{bmatrix} \sum ^{k^a} _{i=1} \binom{|c^a_i|}{2} + \sum ^{k^b} _{j=1} \binom{|c^b_j|}{2} \end{bmatrix}/2 - \begin{bmatrix} \sum ^{k^a} _{i=1} \binom{|c^a_i|}{2} \sum ^{k^b} _{j=1} \binom{|c^b_j|}{2}\end{bmatrix} /\binom{n}{2}} 
    \label{eq:RC}
\end{equation}

\section{Experimental Results}~\label{sec:results}
As described in Section~\ref{sec:goalexpsep}, for every individual in each population, we computed the objective function presented in Section~\ref{sec:obj} and compared their results with the respective values of the true partition to determine their admissibility.

\begin{table}[!htb]
\centering
\scalebox{0.7}{
\begin{tabular}{|l|l|c|c|c|c|c|c|c|c|c|c|c|c|c|c|c|c|c|}
\hline
 {\textbf{G}}& {\rotatebox[x=2.32cm]{360}{\textbf{Datasets}}}&	\multicolumn{1}{l|}{\rot{{$Ent$}}}&	\multicolumn{1}{l|}{\rot{{$Dev$}}}&	\multicolumn{1}{l|}{\rot{{$Var$}}}&	\multicolumn{1}{l|}{\rot{{$TWCV$}}}&	\multicolumn{1}{l|}{\rot{{$CH$}}}&	\multicolumn{1}{l|}{\rot{{DB}}}&	\multicolumn{1}{l|}{\rot{{$Dunn$}}}&	\multicolumn{1}{l|}{\rot{{$Mod$}}}&	\multicolumn{1}{l|}{\rot{\textbf{$Sil$}}}&	\multicolumn{1}{l|}{\rot{\textbf{$PBM$}}}&	\multicolumn{1}{l|}{\rot{\textbf{$XB$}}}&	\multicolumn{1}{l|}{\rot{{$ABGSS$}}}&	\multicolumn{1}{l|}{\rot{{$Sep_{AL}$}}}&	\multicolumn{1}{l|}{\rot{{$Sep_{CL}$}}}&	\multicolumn{1}{l|}{\rot{{$Sep_{graph}$}}}&	\multicolumn{1}{l|}{\rot{{$Con$}}}&	\multicolumn{1}{l|}{\rot{{$DCD$}}}\\\hline	
\multirow{12}{*}{G1}&	\texttt{R15}&	$\times$&	&	&	&	&	&	$\times$&	$\times$&	&	$\times$&	&	$\times$&	$\times$&	&	\checkmark&	$\times$&	$\times$\\\cline{2-19}
&	\texttt{D31}&	$\times$&	&	&	&	&	&	$\times$&	$\times$&	$\times$&	$\times$&	&	$\times$&	$\times$&	&	$\times$&	$\times$&	$\times$\\\cline{2-19}
&	\texttt{Engytime}&	$\times$&	&	&	&	&	$\times$&	$\times$&	$\times$&	$\times$&	$\times$&	&	&	$\times$&	&	$\times$&	$\times$&	$\times$\\\cline{2-19}
&	\texttt{Sizes5}&	$\times$&	&	&	&	&	&	$\times$&	$\times$&	&	&	&	&	&	&	&	$\times$&	$\times$\\\cline{2-19}
&	\texttt{Square1}&	$\times$&	&	&	&	&	&	$\times$&	$\times$&	&	$\times$&	&	&	&	&	&	$\times$&	$\times$\\\cline{2-19}
&	\texttt{Square4}&	$\times$&	&	&	&	&	&	$\times$&	$\times$&	&	$\times$&	&	&	&	&	$\times$&	$\times$&	$\times$\\\cline{2-19}
&	\texttt{Twenty}&	$\times$&	$\times$&	$\times$&	$\times$&	\checkmark&	\checkmark&	\checkmark&	$\times$&	\checkmark&	\checkmark&	\checkmark&	$\times$&	$\times$&	$\times$&	$\times$&	$\times$&	$\times$\\\cline{2-19}
&	\texttt{Fourty}&	$\times$&	$\times$&	$\times$&	$\times$&	\checkmark&	\checkmark&	\checkmark&	$\times$&	\checkmark&	\checkmark&	\checkmark&	$\times$&	$\times$&	$\times$&	\checkmark&	$\times$&	$\times$\\\cline{2-19}
&	\texttt{Sph\_5\_2}&	$\times$&	$\times$&	$\times$&	$\times$&	&	&	$\times$&	$\times$&	&	$\times$&	&	$\times$&	$\times$&	$\times$&	$\times$&	$\times$&	$\times$\\\cline{2-19}
&	\texttt{Sph\_6\_2}&	$\times$&	$\times$&	$\times$&	$\times$&	\checkmark&	\checkmark&	$\times$&	$\times$&	\checkmark&	$\times$&	$\times$&	$\times$&	$\times$&	$\times$&	$\times$&	\checkmark&	$\times$\\\cline{2-19}
&	\texttt{Sph\_9\_2}&	$\times$&	&	&	&	&	&	$\times$&	$\times$&	&	$\times$&	&	$\times$&	$\times$&	&	$\times$&	$\times$&	$\times$\\\cline{2-19}
&	\texttt{Sph\_10\_2}&	$\times$&	$\times$&	$\times$&	$\times$&	&	$\times$&	$\times$&	$\times$&	&	$\times$&	&	$\times$&	$\times$&	$\times$&	$\times$&	$\times$&	$\times$\\\hline
\multirow{3}{*}{G2}&	\texttt{ds2c2sc13\_S1}&	$\times$&	$\times$&	$\times$&	$\times$&	\checkmark&	\checkmark&	\checkmark&	$\times$&	\checkmark&	\checkmark&	\checkmark&	\checkmark&	\checkmark&	$\times$&	\checkmark&	\checkmark&	\checkmark\\\cline{2-19}
&	\texttt{ds2c2sc13\_S2}&	$\times$&	$\times$&	$\times$&	$\times$&	$\times$&	$\times$&	$\times$&	$\times$&	\checkmark&	$\times$&	$\times$&	$\times$&	$\times$&	$\times$&	$\times$&	\checkmark&	$\times$\\\cline{2-19}
&	\texttt{ds2c2sc13\_S3}&	$\times$&	$\times$&	$\times$&	$\times$&	$\times$&	$\times$&	$\times$&	$\times$&	$\times$&	$\times$&	$\times$&	$\times$&	$\times$&	$\times$&	$\times$&	$\times$&	$\times$\\\hline
\multirow{3}{*}{G3}&	\texttt{Long1}&	$\times$&	$\times$&	$\times$&	$\times$&	$\times$&	$\times$&	\checkmark&	\checkmark&	$\times$&	\checkmark&	$\times$&	\checkmark&	$\times$&	$\times$&	\checkmark&	\checkmark&	\checkmark\\\cline{2-19}
&	\texttt{Pat2}&	$\times$&	$\times$&	$\times$&	$\times$&	\checkmark&	$\times$&	\checkmark&	$\times$&	$\times$&	\checkmark&	$\times$&	\checkmark&	$\times$&	$\times$&	$\times$&	\checkmark&	$\times$\\\cline{2-19}
&	\texttt{Spiral}&	$\times$&	$\times$&	$\times$&	$\times$&	\checkmark&	$\times$&	\checkmark&	$\times$&	$\times$&	\checkmark&	\checkmark&	\checkmark&	$\times$&	$\times$&	\checkmark&	\checkmark&	$\times$\\\hline
\multirow{6}{*}{G4}&	\texttt{3MC}&	$\times$&	$\times$&	$\times$&	$\times$&	$\times$&	\checkmark&	\checkmark&	$\times$&	\checkmark&	\checkmark&	\checkmark&	$\times$&	\checkmark&	$\times$&	\checkmark&	\checkmark&	$\times$\\\cline{2-19}
&	\texttt{DS-850}&	$\times$&	&	&	&	&	&	&	$\times$&	&	&	&	&	$\times$&	&	&	$\times$&	$\times$\\\cline{2-19}
&	\texttt{Aggregation}&	$\times$&	&	&	&	&	&	$\times$&	$\times$&	&	$\times$&	&	$\times$&	$\times$&	&	$\times$&	$\times$&	$\times$\\\cline{2-19}
&	\texttt{Complex9}&	$\times$&	$\times$&	$\times$&	$\times$&	$\times$&	$\times$&	$\times$&	$\times$&	$\times$&	$\times$&	$\times$&	$\times$&	$\times$&	$\times$&	$\times$&	$\times$&	$\times$\\\cline{2-19}
&	\texttt{Pat1}&	$\times$&	$\times$&	$\times$&	$\times$&	$\times$&	$\times$&	$\times$&	$\times$&	$\times$&	$\times$&	$\times$&	$\times$&	$\times$&	$\times$&	$\times$&	$\times$&	$\times$\\\cline{2-19}
&	\texttt{Spiralsquare}&	$\times$&	&	&	&	$\times$&	$\times$&	$\times$&	$\times$&	$\times$&	$\times$&	&	$\times$&	$\times$&	&	\checkmark&	$\times$&	$\times$\\\hline
\end{tabular}}
\caption{Results of the analysis of the admissibility of the objective functions considering an initialization with MST-clustering}
	\label{tab:obj_iniPopMST}
\end{table} 

Table \ref{tab:obj_iniPopMST} and Table~\ref{tab:obj_iniPopKM} present the detailed results of the admissibility in terms of the initialization with MST-clustering and KM, respectively. In these tables, we point out the objective functions that are inadmissible ($\times$) for each dataset. Also, in terms of the potentially admissible objective function, we consider two other classes: (i) the true partition was found in the initial population,  where optimization is not required (\checkmark), and (ii) the objective function is admissible and optimizing should be performed to find the optimal solutions (blank cells). For a complementary view, \ref{app1} presents the box-plot of the ARI results of each initialization algorithm.

\begin{table}[!ht]
\centering
\scalebox{0.7}{
\begin{tabular}{|l|l|c|c|c|c|c|c|c|c|c|c|c|c|c|c|c|c|c|}\hline
{\textbf{G}}& {\rotatebox[x=2.32cm]{360}{\textbf{Datasets}}}&	\multicolumn{1}{l|}{\rot{{$Ent$}}}&	\multicolumn{1}{l|}{\rot{{$Dev$}}}&	\multicolumn{1}{l|}{\rot{{$Var$}}}&	\multicolumn{1}{l|}{\rot{{$TWCV$}}}&	\multicolumn{1}{l|}{\rot{{$CH$}}}&	\multicolumn{1}{l|}{\rot{{DB}}}&	\multicolumn{1}{l|}{\rot{{$Dunn$}}}&	\multicolumn{1}{l|}{\rot{{$Mod$}}}&	\multicolumn{1}{l|}{\rot{\textbf{$Sil$}}}&	\multicolumn{1}{l|}{\rot{\textbf{$PBM$}}}&	\multicolumn{1}{l|}{\rot{\textbf{$XB$}}}&	\multicolumn{1}{l|}{\rot{{$ABGSS$}}}&	\multicolumn{1}{l|}{\rot{{$Sep_{AL}$}}}&	\multicolumn{1}{l|}{\rot{{$Sep_{CL}$}}}&	\multicolumn{1}{l|}{\rot{{$Sep_{graph}$}}}&	\multicolumn{1}{l|}{\rot{{$Con$}}}&	\multicolumn{1}{l|}{\rot{{$DCD$}}}\\\hline	
\multirow{12}{*}{G1}&	\texttt{R15}&	$\times$&	$\times$&	$\times$&	$\times$&	$\times$&	$\times$&	$\times$&	$\times$&	$\times$&	$\times$&	$\times$&	$\times$&	$\times$&	$\times$&	$\times$&	$\times$&	$\times$\\\cline{2-19}
&	\texttt{D31}&	$\times$&	$\times$&	$\times$&	$\times$&	$\times$&	$\times$&	$\times$&	$\times$&	$\times$&	$\times$&	$\times$&	$\times$&	$\times$&	$\times$&	$\times$&	$\times$&	$\times$\\\cline{2-19}
&	\texttt{Engytime}&	$\times$&	$\times$&	$\times$&	$\times$&	$\times$&	$\times$&	$\times$&	$\times$&	$\times$&	$\times$&	$\times$&	$\times$&	$\times$&	$\times$&	&	$\times$&	\\\cline{2-19}
&	\texttt{Sizes5}&	$\times$&	$\times$&	$\times$&	$\times$&	$\times$&	&	&	$\times$&	$\times$&	&	&	$\times$&	&	$\times$&	&	&	$\times$\\\cline{2-19}
&	\texttt{Square1}&	$\times$&	$\times$&	$\times$&	$\times$&	$\times$&	$\times$&	$\times$&	$\times$&	$\times$&	$\times$&	$\times$&	$\times$&	$\times$&	$\times$&	$\times$&	$\times$&	$\times$\\\cline{2-19}
&	\texttt{Square4}&	$\times$&	$\times$&	$\times$&	$\times$&	$\times$&	$\times$&	$\times$&	$\times$&	$\times$&	$\times$&	$\times$&	$\times$&	$\times$&	$\times$&	$\times$&	$\times$&	$\times$\\\cline{2-19}
&	\texttt{Twenty}&	$\times$&	$\times$&	$\times$&	$\times$&	\checkmark&	\checkmark&	\checkmark&	$\times$&	\checkmark&	\checkmark&	\checkmark&	$\times$&	$\times$&	$\times$&	$\times$&	\checkmark&	$\times$\\\cline{2-19}
&	\texttt{Fourty}&	$\times$&	$\times$&	$\times$&	$\times$&	\checkmark&	\checkmark&	\checkmark&	$\times$&	\checkmark&	\checkmark&	\checkmark&	$\times$&	$\times$&	$\times$&	\checkmark&	$\times$&	$\times$\\\cline{2-19}
&	\texttt{Sph\_5\_2}&	$\times$&	$\times$&	$\times$&	$\times$&	$\times$&	$\times$&	$\times$&	$\times$&	$\times$&	$\times$&	$\times$&	$\times$&	$\times$&	$\times$&	$\times$&	$\times$&	$\times$\\\cline{2-19}
&	\texttt{Sph\_6\_2}&	$\times$&	$\times$&	$\times$&	$\times$&	\checkmark&	\checkmark&	$\times$&	$\times$&	\checkmark&	$\times$&	$\times$&	$\times$&	$\times$&	$\times$&	$\times$&	\checkmark&	$\times$\\\cline{2-19}
&	\texttt{Sph\_9\_2}&	$\times$&	$\times$&	$\times$&	$\times$&	$\times$&	$\times$&	$\times$&	$\times$&	$\times$&	$\times$&	$\times$&	$\times$&	$\times$&	$\times$&	$\times$&	$\times$&	$\times$\\\cline{2-19}
&	\texttt{Sph\_10\_2}&	$\times$&	$\times$&	$\times$&	$\times$&	$\times$&	$\times$&	$\times$&	$\times$&	&	$\times$&	$\times$&	$\times$&	$\times$&	$\times$&	$\times$&	$\times$&	$\times$\\\hline
\multirow{3}{*}{G2}&	\texttt{ds2c2sc13\_S1}&	\checkmark&	$\times$&	$\times$&	$\times$&	\checkmark&	\checkmark&	\checkmark&	$\times$&	\checkmark&	\checkmark&	\checkmark&	\checkmark&	\checkmark&	$\times$&	&	\checkmark&	\checkmark\\\cline{2-19}
&	\texttt{ds2c2sc13\_S2}&	$\times$&	$\times$&	$\times$&	$\times$&	$\times$&	$\times$&	$\times$&	$\times$&	$\times$&	$\times$&	$\times$&	$\times$&	$\times$&	$\times$&	$\times$&	\checkmark&	$\times$\\\cline{2-19}
&	\texttt{ds2c2sc13\_S3}&	$\times$&	$\times$&	$\times$&	$\times$&	$\times$&	$\times$&	$\times$&	$\times$&	$\times$&	$\times$&	$\times$&	$\times$&	$\times$&	$\times$&	$\times$&	$\times$&	$\times$\\\hline
\multirow{3}{*}{G3}&	\texttt{Long1}&	$\times$&	$\times$&	$\times$&	$\times$&	$\times$&	$\times$&	&	&	$\times$&	&	$\times$&	$\times$&	$\times$&	$\times$&	&	&	$\times$\\\cline{2-19}
&	\texttt{Pat2}&	$\times$&	$\times$&	$\times$&	$\times$&	$\times$&	$\times$&	$\times$&	&	$\times$&	$\times$&	$\times$&	$\times$&	$\times$&	$\times$&	&	&	$\times$\\\cline{2-19}
&	\texttt{Spiral}&	$\times$&	$\times$&	$\times$&	$\times$&	$\times$&	$\times$&	&	&	$\times$&	&	$\times$&	$\times$&	$\times$&	$\times$&	&	&	$\times$\\\hline
\multirow{6}{*}{G4}&	\texttt{3MC}&	$\times$&	$\times$&	$\times$&	$\times$&	$\times$&	$\times$&	&	$\times$&	&	&	$\times$&	$\times$&	$\times$&	$\times$&	$\times$&	&	$\times$\\\cline{2-19}
&	\texttt{DS-850}&	$\times$&	$\times$&	$\times$&	$\times$&	$\times$&	&	&	$\times$&	&	&	$\times$&	$\times$&	$\times$&	$\times$&	$\times$&	&	$\times$\\\cline{2-19}
&	\texttt{Aggregation}&	$\times$&	$\times$&	$\times$&	$\times$&	$\times$&	&	$\times$&	$\times$&	$\times$&	$\times$&	$\times$&	$\times$&	$\times$&	$\times$&	$\times$&	&	$\times$\\\cline{2-19}
&	\texttt{Complex9}&	$\times$&	$\times$&	$\times$&	$\times$&	$\times$&	$\times$&	&	$\times$&	$\times$&	&	$\times$&	$\times$&	$\times$&	$\times$&	$\times$&	&	$\times$\\\cline{2-19}
&	\texttt{Pat1}&	$\times$&	$\times$&	$\times$&	$\times$&	$\times$&	$\times$&	$\times$&	&	$\times$&	$\times$&	$\times$&	$\times$&	$\times$&	$\times$&	$\times$&	&	$\times$\\\cline{2-19}
&	\texttt{Spiralsquare}&	$\times$&	$\times$&	$\times$&	$\times$&	$\times$&	$\times$&	&	$\times$&	$\times$&	&	$\times$&	$\times$&	$\times$&	$\times$&	&	&	$\times$\\\hline
\end{tabular}}
\caption{Results of the analysis of the admissibility of the objective functions considering an initialization with KM}
	\label{tab:obj_iniPopKM}
\end{table} 

Table~\ref{tab:obj_iniPop} summarizes the results of all initializations. Since the initialization of the AL, SL, and SNN are comparable with the results of the MST-clustering and KM, we compiled the results by counting the number of datasets in which the objective functions are inadmissible. Column \textbf{IN} denotes the total number of the datasets where each objective function is inadmissible, and column \textbf{OP} refers to the total number of the datasets where the optimal solution is provided in the initial population. For example, the fields \textbf{IN} fulfilled with 24 mean that a specific objective function is inadmissible for any of the analyzed datasets.

\begin{table}[!htb]
\centering
\scalebox{0.7}{
\begin{tabular}{|l|l|rr|rr|rr|rr|rr|}
\hline
\multicolumn{1}{|c|}{\multirow{2}{*}{\textbf{Type}}} & \multicolumn{1}{c|}{\multirow{2}{*}{\textbf{Objectives}}} & \multicolumn{2}{c|}{\textbf{MST}} & \multicolumn{2}{c|}{\textbf{SNN}} & \multicolumn{2}{c|}{\textbf{SL}} & \multicolumn{2}{c|}{\textbf{AL}} & \multicolumn{2}{c|}{\textbf{KM}} \\ \cline{3-12} 
\multicolumn{1}{|c|}{} & \multicolumn{1}{c|}{} & \multicolumn{1}{c|}{\textbf{IN}} & \multicolumn{1}{c|}{\textbf{OP}} & \multicolumn{1}{c|}{\textbf{IN}} & \multicolumn{1}{c|}{\textbf{OP}} & \multicolumn{1}{c|}{\textbf{IN}} & \multicolumn{1}{c|}{\textbf{OP}} & \multicolumn{1}{c|}{\textbf{IN}} & \multicolumn{1}{c|}{\textbf{OP}} & \multicolumn{1}{c|}{\textbf{IN}} & \multicolumn{1}{c|}{\textbf{OP}} \\ \hline
\multirow{4}{*}{\textbf{Compactness}} & {$Ent$} & \multicolumn{1}{r|}{24} & - & \multicolumn{1}{r|}{20} & 3 & \multicolumn{1}{r|}{24} & - & \multicolumn{1}{r|}{23} & 1 & \multicolumn{1}{r|}{23} & 1 \\ \cline{2-12} 
 & $Dev$ & \multicolumn{1}{r|}{14} & - & \multicolumn{1}{r|}{11} & 3 & \multicolumn{1}{r|}{12} & - & \multicolumn{1}{r|}{24} & - & \multicolumn{1}{r|}{24} & - \\ \cline{2-12} 
 & $Var$ & \multicolumn{1}{r|}{14} & - & \multicolumn{1}{r|}{11} & 3 & \multicolumn{1}{r|}{12} & - & \multicolumn{1}{r|}{24} & - & \multicolumn{1}{r|}{24} & - \\ \cline{2-12} 
 & $TWCV$ & \multicolumn{1}{r|}{14} & - & \multicolumn{1}{r|}{17} & 1 & \multicolumn{1}{r|}{15} & - & \multicolumn{1}{r|}{24} & - & \multicolumn{1}{r|}{24} & - \\ \hline
\multirow{7}{*}{\textbf{\begin{tabular}[c]{@{}l@{}}Compactness\\ and\\ Separation\end{tabular}}} & {$CH$} & \multicolumn{1}{r|}{7} & 6 & \multicolumn{1}{r|}{5} & 9 & \multicolumn{1}{r|}{4} & 8 & \multicolumn{1}{r|}{13} & 3 & \multicolumn{1}{r|}{20} & 4 \\ \cline{2-12} 
 & {$DB$} & \multicolumn{1}{r|}{10} & 5 & \multicolumn{1}{r|}{9} & 8 & \multicolumn{1}{r|}{14} & 4 & \multicolumn{1}{r|}{18} & 4 & \multicolumn{1}{r|}{17} & 4 \\ \cline{2-12} 
 & {$Dunn$} & \multicolumn{1}{r|}{16} & 7 & \multicolumn{1}{r|}{16} & 7 & \multicolumn{1}{r|}{17} & 6 & \multicolumn{1}{r|}{16} & 4 & \multicolumn{1}{r|}{14} & 3 \\ \cline{2-12} 
 & {$Mod$} & \multicolumn{1}{r|}{23} & 1 & \multicolumn{1}{r|}{22} & 2 & \multicolumn{1}{r|}{21} & 1 & \multicolumn{1}{r|}{22} & - & \multicolumn{1}{r|}{20} & - \\ \cline{2-12} 
 & {$Sil$} & \multicolumn{1}{r|}{7} & 5 & \multicolumn{1}{r|}{6} & 7 & \multicolumn{1}{r|}{7} & 6 & \multicolumn{1}{r|}{14} & 3 & \multicolumn{1}{r|}{20} & 3 \\ \cline{2-12} 
 
& {$PBM$} & \multicolumn{1}{r|}{9} & 6 & \multicolumn{1}{r|}{5} & 10 & \multicolumn{1}{r|}{7} & 8 & \multicolumn{1}{r|}{18} & 4 & \multicolumn{1}{r|}{17} & 4 \\ \cline{2-12} 
& {$XB$} & \multicolumn{1}{r|}{15} & 7 & \multicolumn{1}{r|}{16} & 7 & \multicolumn{1}{r|}{17} & 6 & \multicolumn{1}{r|}{15} & 4 & \multicolumn{1}{r|}{14} & 3 \\ \hline
\multirow{4}{*}{\textbf{Separation}} & {$ABSS$} & \multicolumn{1}{r|}{16} & 3 & \multicolumn{1}{r|}{13} & 5 & \multicolumn{1}{r|}{13} & 4 & \multicolumn{1}{r|}{21} & 1 & \multicolumn{1}{r|}{23} & 1 \\ \cline{2-12} 
 & {$Sep_{AL}$} & \multicolumn{1}{r|}{19} & 2 & \multicolumn{1}{r|}{13} & 5 & \multicolumn{1}{r|}{20} & 3 & \multicolumn{1}{r|}{23} & 1 & \multicolumn{1}{r|}{22} & 1 \\ \cline{2-12} 
 & {$Sep_{CL}$} & \multicolumn{1}{r|}{14} & - & \multicolumn{1}{r|}{11} & 3 & \multicolumn{1}{r|}{12} & - & \multicolumn{1}{r|}{24} & - & \multicolumn{1}{r|}{24} & - \\ \cline{2-12} 
 & {$Sep_{graph}$} & \multicolumn{1}{r|}{14} & 7 & \multicolumn{1}{r|}{6} & 11 & \multicolumn{1}{r|}{13} & 4 & \multicolumn{1}{r|}{19} & 1 & \multicolumn{1}{r|}{16} & 1 \\ \hline
\multirow{2}{*}{\textbf{Connectedness}} & {$Con$} & \multicolumn{1}{r|}{17} & 7 & \multicolumn{1}{r|}{17} & 7 & \multicolumn{1}{r|}{17} & 6 & \multicolumn{1}{r|}{15} & 4 & \multicolumn{1}{r|}{10} & 4 \\ \cline{2-12} 
 & {$DCD$} & \multicolumn{1}{r|}{22} & 2 & \multicolumn{1}{r|}{24} & - & \multicolumn{1}{r|}{24} & - & \multicolumn{1}{r|}{24} & - & \multicolumn{1}{r|}{22} & 1 \\ \hline
\end{tabular}}
\caption{A summary  of the results regarding the analysis of the admissibility of the objective functions}\label{tab:obj_iniPop}
\end{table}

\section{Discussion}~\label{sec:disc}
By analyzing Table \ref{tab:obj_iniPopMST}, we observed that every connectedness criterion ($Con$ and $DCD$) provides results in which there is no space to optimize in a such direction since they are inadmissible, or the optimal result was found in the initial population (cells marked with $\times$ and \checkmark). In contrast, some objective functions that take into account the compactness or/and separation criteria could be used in the optimization of the datasets that include the Gaussian-like clusters and hyper-spherical clusters that have some degree of  overlap in G1, and heterogeneous data structures with close objects between the clusters in G4  (blank cells). In general, MST-clustering fails in detecting close or overlapping clusters, and the use of a complementary objective function that considers a search in different criteria (direction) covered in the initialization can lead the EMOC to obtain better results.  

In terms of the results shown in Table 4, we observe that for most of the objective functions, there is no space for optimization (cells marked with \checkmark or $\times$) when KM is applied in the initialization. Only the objective functions $Dunn$, $PBM$, $Sep_{graph}$, and $Con$ have at least five datasets in which it is possible to improve the results (blank cells). In particular, we observed that $Con$ has space to be optimized in ten datasets, including all datasets present in G3 and G4. In this case, KM fails to detect the elongated clusters, and the use of $Con$ could lead the EMOC to find this kind of structure present in G3 and G4. 
In general, these results point out that the initializations with KM and AL provide limitations in optimizing most of the compactness or separation criteria for most datasets. 
 
In the initializations with MST-clustering, SNN, and SL, the objective functions present similar behavior.  All objective functions are inadmissible or produce the best results for datasets with well-separated clusters. Consequently, there is no space for optimizing any evaluated criteria for these features (well-separated clusters and initialization with MTS, SNN, and SL). 

In terms of the EMOCs, we verified an issue in the design of the approaches that consider the same clustering criteria in the initialization strategy and the objective functions, in which the evolutionary optimization could not be adequate. 
For example, in \cite{Handl2005b}, it is applied the KM in the initialization and the pair of objectives ($Var$ and $Dev$). In this case, the initial population has solutions that  either reach the optimal results or exceed the boundaries of feasible search space to find compacted clusters. Therefore, optimization in this direction would not be necessary. 
Furthermore, these objective functions are very similar in their formulation, which limits the capabilities of the EMOC in generating a diverse set of solutions. 
MOCLE, beyond other EMOCs, also presents a similar design, in which every objective function is inadmissible for all the datasets in terms of at least one method used in the initialization. 

It is important to note that, in general, EMOCs present in the literature do not use restrictions as defined in  Eq.~\ref{mop_r1} (see Section \ref{section:background}). They usually apply the restrictions as objective functions to maintain good solutions or restrict the search in some direction. This case is different from the above scenario, in which the initialization strategy limits the search to all the objective functions used in the multi-objective approach. 

These results, presented in Table~\ref{tab:obj_iniPop}, show that for every analyzed initialization, there is no objective function that is admissible in all the datasets, answering our first research question.
In the following sub-section, we analyzed the clustering results considering the optimization of the selected objective functions, extending our analysis. We picked the objective functions that presented the lowest results of the inadmissibility considering the initialization with MTS-clustering. 

\subsection{Analysis of the objective functions in the optimization}
Aiming to answer the second research question, we analyzed one initialization strategy, considering different scenarios of  the combination of objective functions. 
In particular,  we analyze the behavior of the objective functions in order to improve the detection of no well-separated clusters and close clusters in the heterogeneous data structures in terms of the results presented in Table \ref{tab:obj_iniPopMST}.  Hence, we selected one objective function per criterion that presented the lowest number of datasets in which they are inadmissible: $Var$, $CH$, $Sep_{CL}$, and $Con$. Moreover, as described in Section~\ref{sec:experiment}, $\Delta$-MOCK was chosen because it is a recent approach based on an established EMOC that provides features that allow us to explore the use of MST-clustering in the initialization. 

It should be noted that in this paper we demonstrate how to perform the analysis of the objective functions while considering a particular EMOC and specific goals. Different scenarios, considering other initializations (or even other EMOCs), can lead to different admissibility results and different clustering performance (ARI).  

Table \ref{tab:obj_iniPopMST2obj} presents the average ARI and standard deviation of 30 runs for each dataset generated by the $\Delta$-MOCK considering different combinations of the selected objective function. For a complementary view, \ref{app2} presents the box-plot of the ARI results for each pair of objective functions. The MST column refers to the best ARI found in the partitions generated by the MST-clustering. The underlined results point out the objective functions in which the optimization generated solutions that dominate the true partition.  

\afterpage{
\begin{landscape}
\begin{table}[!ht]
\scalebox{0.68}{
\begin{tabular}{|cl|r|rl|rl|rl|rl|rl|rl|}
\hline
\multicolumn{1}{|c|}{\textbf{G}} & \multicolumn{1}{c|}{\textbf{Dataset}} & \multicolumn{1}{c|}{{MST}} & \multicolumn{2}{c|}{\textbf{($Var$, $Sep_{CL}$)}} & \multicolumn{2}{c|}{\textbf{($Ch$, $Sep_{CL}$)}} & \multicolumn{2}{c|}{\textbf{($Var$, $CH$)}} & \multicolumn{2}{c|}{\textbf{($CH$, $Con$)}} & \multicolumn{2}{c|}{\textbf{($Var$, $Con$)}} & \multicolumn{2}{c|}{\textbf{($Con$, $Sep_{CL}$)}} \\ \hline
\multicolumn{1}{|c|}{\multirow{12}{*}{G1 }} & \texttt{R15} & 0.7203 & {\underline{0.4312}} & $\pm$ 1.58E-02 & {\underline{0.9844}} & $\pm$ 8.49E-03 & {\underline{0.9857}} & $\pm$ 1.60E-03 & {\underline{\textbf{0.9928}}} & $\pm$ 7.90E-16 & {\underline{0.9914}} & $\pm$ 9.62E-02 & {\underline{0.9892}} & $\pm$ 5.11E-03 \\ \cline{2-15} 
\multicolumn{1}{|c|}{} & \texttt{D31} & 0.4708 & {\underline{0.5515}} & $\pm$ 1.86E-02 & {0.7358} & $\pm$ 3.13E-02 & {\textbf{0.7620}} & $\pm$ 1.60E-03 & {0.4327} & $\pm$ 8.95E-02 & {0.7455} & $\pm$ 2.15E-02 & {0.7383} & $\pm$ 1.81E-02 \\ \cline{2-15} 
\multicolumn{1}{|c|}{} & \texttt{Engytime} & 0.0076 & {\underline{0.0369}} & $\pm$ 2.50E-03 & {\underline{0.4945}} & $\pm$ 1.12E-01 & {\underline{0.4849}} & $\pm$ 1.60E-03 & {\underline{0.7638}} & $\pm$ 8.95E-02 & {\underline{0.7847}} & $\pm$ 3.39E-02 & {\underline{\textbf{0.8247}}} & $\pm$ 2.00E-02 \\ \cline{2-15} 
\multicolumn{1}{|c|}{} & \texttt{Sizes5} & 0.5642 & {\underline{0.0333}} & $\pm$ 1.26E-03 & {\underline{0.3464}} & $\pm$ 1.16E-01 & {\underline{0.3027}} & $\pm$ 6.52E-02 & {\textbf{0.9638}} & $\pm$ 3.27E-03 & {\underline{0.9622}} & $\pm$ 7.63E-03 & {\underline{0.9488}} & $\pm$ 1.35E-02 \\ \cline{2-15} 
\multicolumn{1}{|c|}{} & \texttt{Square1} & 0.3711 & {\underline{0.1315}} & $\pm$ 6.13E-03 & {0.9457} & $\pm$ 1.73E-02 & {0.9504} & $\pm$ 1.74E-02 & {\underline{\textbf{0.9761}}} & $\pm$ 3.39E-16 & {\underline{0.9728}} & $\pm$ 4.82E-03 & {\underline{0.9714}} & $\pm$ 6.46E-03 \\ \cline{2-15} 
\multicolumn{1}{|c|}{} & \texttt{Square4} & 0.4570 & {\underline{0.1389}} & $\pm$ 4.60E-03 & {0.7282} & $\pm$ 6.85E-02 & {0.7101} & $\pm$ 7.78E-02 & {\underline{\textbf{0.7847}}} & $\pm$ 2.05E-02 & {\underline{0.7739}} & $\pm$ 4.82E-03 & {\underline{0.7678}} & $\pm$ 2.56E-02 \\ \cline{2-15} 
\multicolumn{1}{|c|}{} & \texttt{Twenty} & \textbf{1.0000} & {\underline{0.4496}} & $\pm$ 4.60E-03 & {\textbf{1.0000}} & $\pm$ 0.00E+00 & {\textbf{1.0000}} & $\pm$ 0.00E+00 & {\textbf{1.0000}} & $\pm$ 0.00E+00 & {\textbf{1.0000}} & $\pm$ 0.00E+00 & {\textbf{1.0000}} & $\pm$ 0.00E+00 \\ \cline{2-15} 
\multicolumn{1}{|c|}{} & \texttt{Fourty} & \textbf{1.0000} & {\underline{0.6833}} & $\pm$ 1.27E-02 & {\textbf{1.0000}} & $\pm$ 0.00E+00 & {\textbf{1.0000}} & $\pm$ 0.00E+00 & {\textbf{1.0000}} & $\pm$ 0.00E+00 & {\textbf{1.0000}} & $\pm$ 0.00E+00 & {\textbf{1.0000}} & $\pm$ 0.00E+00 \\ \cline{2-15} 
\multicolumn{1}{|c|}{} & \texttt{Sph\_5\_2} & 0.7315 & {\underline{0.1325}} & $\pm$ 5.45E-03 & {\textbf{0.9239}} & $\pm$ 3.95E-02 & {0.9072} & $\pm$ 1.37E-01 & {0.9205} & $\pm$ 2.58E-02 & {\underline{0.9019}} & $\pm$ 2.15E-02 & {\underline{0.8932}} & $\pm$ 3.72E-02 \\ \cline{2-15} 
\multicolumn{1}{|c|}{} & \texttt{Sph\_6\_2} & \textbf{1.0000} & {\underline{0.1665}} & $\pm$ 6.86E-03 & {\textbf{1.0000}} & $\pm$ 0.00E+00 & {\textbf{1.0000}} & $\pm$ 0.00E+00 & {\textbf{1.0000}} & $\pm$ 0.00E+00 & {\textbf{1.0000}} & $\pm$ 0.00E+00 & {\textbf{1.0000}} & $\pm$ 0.00E+00 \\ \cline{2-15} 
\multicolumn{1}{|c|}{} & \texttt{Sph\_9\_2} & 0.3057 & {\underline{0.2396}} & $\pm$ 7.92E-03 & {0.7079} & $\pm$ 4.70E-02 & {0.7230} & $\pm$ 4.00E-02 & {0.5799} & $\pm$ 1.48E-01 & {\underline{\textbf{0.7228}}} & $\pm$ 2.25E-02 & {\underline{0.7110}} & $\pm$ 2.85E-02 \\ \cline{2-15} 
\multicolumn{1}{|c|}{} & \texttt{Sph\_10\_2} & 0.8651 & {\underline{0.2765}} & $\pm$ 8.36E-03 & {0.7008} & $\pm$ 2.69E-01 & {0.8313} & $\pm$ 2.06E-01 & {\textbf{0.9857}} & $\pm$ 3.85E-03 & {\underline{0.9743}} & $\pm$ 1.08E-02 & {\underline{0.9732}} & $\pm$ 8.89E-03 \\ \hline
\multicolumn{1}{|c|}{\multirow{3}{*}{G2}} & \texttt{ds2c2sc13\_S1} & \textbf{1.0000} & {\underline{0.0361}} & $\pm$ 1.71E-03 & {\underline{0.0622}} & $\pm$ 7.16E-03 & {\underline{0.0707}} & $\pm$ 5.08E-02 & {\textbf{1.0000}} & $\pm$ 0.00E+00 & {\underline{0.3520}} & $\pm$ 2.82E-16 & {\underline{0.3520}} & $\pm$ 2.82E-16 \\ \cline{2-15} 
\multicolumn{1}{|c|}{} & \texttt{ds2c2sc13\_S2} & \textbf{1.0000} & {\underline{0.1444}} & $\pm$ 6.64E-03 & {\underline{0.2323}} & $\pm$ 2.39E-02 & {\underline{0.2409}} & $\pm$ 3.05E-02 & {\underline{0.8455}} & $\pm$ 1.60E-01 & {\underline{0.9518}} & $\pm$ 5.65E-16 & {\underline{0.9518}} & $\pm$ 5.65E-16 \\ \cline{2-15} 
\multicolumn{1}{|c|}{} & \texttt{ds2c2sc13\_S3} & \textbf{0.9952} & {\underline{0.3043}} & $\pm$ 8.35E-03 & {\underline{0.4517}} & $\pm$ 3.89E-02 & {\underline{0.4623}} & $\pm$ 3.90E-02 & {\underline{0.5672}} & $\pm$ 1.37E-02 & {\underline{0.8713}} & $\pm$ 4.31E-03 & {\underline{0.9139}} & $\pm$ 1.19E-02 \\ \hline
\multicolumn{1}{|c|}{\multirow{3}{*}{G3}} & \texttt{Long1} & \textbf{1.0000} & {\underline{0.0426}} & $\pm$ 1.89E-03 & {\underline{0.0850}} & $\pm$ 1.28E-02 & {\underline{0.0870}} & $\pm$ 1.25E-02 & {\textbf{1.0000}} & $\pm$ 0.00E+00 & {\textbf{1.0000}} & $\pm$ 0.00E+00 & {\textbf{1.0000}} & $\pm$ 0.00E+00 \\ \cline{2-15} 
\multicolumn{1}{|c|}{} & \texttt{Pat2} & \textbf{1.0000} & {\underline{0.0334}} & $\pm$ 1.64E-03 & {\underline{0.0638}} & $\pm$ 4.13E-03 & {\underline{0.0642}} & $\pm$ 6.17E-03 & {\textbf{1.0000}} & $\pm$ 0.00E+00 & {\textbf{1.0000}} & $\pm$ 0.00E+00 & {\textbf{1.0000}} & $\pm$ 0.00E+00 \\ \cline{2-15} 
\multicolumn{1}{|c|}{} & \texttt{Spiral} & \textbf{1.0000} & {\underline{0.7122}} & $\pm$ 9.11E-04 & {\underline{0.7363}} & $\pm$ 2.83E-03 & {\underline{0.7363}} & $\pm$ 2.22E-03 & {\textbf{1.0000}} & $\pm$ 0.00E+00 & {\textbf{1.0000}} & $\pm$ 0.00E+00 & {\textbf{1.0000}} & $\pm$ 0.00E+00 \\ \hline
\multicolumn{1}{|c|}{\multirow{6}{*}{G4}} & \texttt{3MC} & \textbf{1.0000} & {\underline{0.0764}} & $\pm$ 1.80E-03 & {\underline{0.1268}} & $\pm$ 1.37E-02 & {\underline{0.1263}} & $\pm$ 1.42E-02 & {\textbf{1.0000}} & $\pm$ 0.00E+00 & {\textbf{1.0000}} & $\pm$ 0.00E+00 & {\textbf{1.0000}} & $\pm$ 0.00E+00 \\ \cline{2-15} 
\multicolumn{1}{|c|}{} & \texttt{DS-850} & 0.4503 & {\underline{0.1722}} & $\pm$ 7.00E-03 & {\underline{0.8059}} & $\pm$ 7.75E-02 & {\underline{0.7590}} & $\pm$ 1.61E-01 & {\textbf{1.0000}} & $\pm$ 0.00E+00 & {0.9994} & $\pm$ 1.97E-03 & {0.9990} & $\pm$ 2.88E-03 \\ \cline{2-15} 
\multicolumn{1}{|c|}{} & \texttt{Aggregation} & 0.8089 & {\underline{0.1395}} & $\pm$ 5.50E-03 & {\underline{0.5008}} & $\pm$ 5.61E-02 & {\underline{0.4812}} & $\pm$ 8.18E-02 & {0.7722} & $\pm$ 1.05E-01 & {\textbf{0.9388}} & $\pm$ 1.79E-02 & {0.9302} & $\pm$ 2.01E-02 \\ \cline{2-15} 
\multicolumn{1}{|c|}{} & \texttt{Complex9} & 0.9314 & {\underline{0.1308}} & $\pm$ 4.81E-03 & {\underline{0.2186}} & $\pm$ 2.21E-02 & {\underline{0.2151}} & $\pm$ 2.72E-02 & {\underline{0.6262}} & $\pm$ 2.91E-03 & {0.9660} & $\pm$ 3.47E-02 & {\textbf{0.9707}} & $\pm$ 3.33E-02 \\ \cline{2-15} 
\multicolumn{1}{|c|}{} & \texttt{Pat1} & \textbf{1.0000} & {\underline{0.0153}} & $\pm$ 8.23E-04 & {\underline{0.0564}} & $\pm$ 1.23E-02 & {\underline{0.0595}} & $\pm$ 1.90E-02 & {\underline{0.7438}} & $\pm$ 2.26E-16 & {\textbf{1.0000}} & $\pm$ 0.00E+00 & {\textbf{1.0000}} & $\pm$ 0.00E+00 \\ \cline{2-15} 
\multicolumn{1}{|c|}{} & \texttt{Spiralsquare} & 0.9290 & {\underline{0.9292}} & $\pm$ 3.39E-04 & {0.9978} & $\pm$ 1.60E-03 & {\textbf{0.9983}} & $\pm$ 1.09E-03 & {0.7744} & $\pm$ 9.62E-02 & {\underline{0.9980}} & $\pm$ 1.18E-03 & {\underline{0.9976}} & $\pm$ 1.56E-03 \\ \hline
\multicolumn{2}{|l|}{\textbf{MEAN}} & \multicolumn{1}{l|}{\textbf{0.7753}} & \multicolumn{2}{l|}{\textbf{0.2503}} & \multicolumn{2}{l|}{\textbf{0.5794}} & \multicolumn{2}{l|}{\textbf{0.5816}} & \multicolumn{2}{l|}{\textbf{0.8637}} & \multicolumn{2}{l|}{\textbf{0.9128}} & \multicolumn{2}{l|}{\textbf{0.9139}} \\ \hline
\end{tabular}}
\caption{Average ARI and standard deviation of different pairs of objective functions in $\Delta$-MOCK}
\label{tab:obj_iniPopMST2obj}
\end{table}
\end{landscape}}

The results point out that, in general, the use of the pairs of objective functions ($Var$, $Sep_{CL}$), ($Ch$, $Sep_{CL}$), and ($Var$, $CH$) does not provide reliable results, because they lose the relation of connectedness in the solutions when it is not applied any restriction. Besides that, these pairs of objective functions dominate the true partition in most of the datasets. 

In contrast, the pairs of the objectives ($CH$, $Con$), ($Sep_{CL}$, $Con$) and ($Var$, $Con$) provide the mean ARI of all datasets above 0.85, as  shown in the row Mean in Table \ref{tab:obj_iniPopMST2obj}. 
The use of $Con$ as objective function preserves the continuity property of clusters and restricts the search to  providing solutions that correspond to the trade-off between this objective and the other objectives ($CH$, $Var$, or $Sep_{CL}$). The best results are provided by the pairs ($Sep_{CL}$, $Con$) and ($Var$, $Con$), both with a mean ARI above 0.91. 

In general, the results relating to the use of ($Sep_{CL}$, $Con$) and ($Var$, $Con$) show that the ARI was improved in the no well-separated clusters present in G1 and heterogeneous data structures in G4. Besides that, most of the good solutions found in the initialization were preserved in most datasets. However, we can observe a loss of the ARI for the datasets in G2 when compared with the initial population (MST column).  
As shown in Table \ref{tab:obj_iniPopMST}, the objective functions ($Sep_{CL}$, $Con$) or ($Var$, $Con$) were inadmissible ($\times$) or obtained the optimal result (\checkmark) for the datasets in G2 and G3, thus the optimization is not required.    
In this context, the results in Table \ref{tab:obj_iniPopMST2obj} confirm our previous results, in which the optimization of these objective functions could only provide a general cost, and it did not afford any improvement in the clustering results. Moreover, the optimization of these objective functions could worsen the clustering performance, as observed in the G2. 
Furthermore, the optimization of these objective functions has another issue, the domination of the true partitions in most of the datasets in the G1 and G2. In G1, it occurred mainly in the datasets with overlapping clusters. In this case, the size of the neighborhood used in the computation of the $Con$ and the distribution of the points in the boundaries of the overlapping clusters may determine the domination of the true partition. In particular, $Con$ computes the continuity of the data based on the neighborhood; however, an overlapping region might have several nearest neighbors in common, making it difficult to determine which cluster each point in the boundaries belongs to.
Regarding G2, as reported in \cite{kultzak2021multi}, the optimization of the dataset {\tt ds2c2sc13} in $\Delta$-MOCK can produce several solutions with optimal $Con$; as a consequence, for these solutions, the decisions around the evolutionary multi-objective optimization will be taken essentially based on the other criteria, in which the true partition is dominated.

 In summary, we observe that the initialization strategy should be correlated with the restrictions applied in the EMOCs. For example, in $\Delta$-MOCK, the objective function $Con$ takes on this role in order to maintain the high quality clusters found in the initialization. Furthermore, optimizing some groups of datasets is not required because the initialization provides the optimal result. However, in $\Delta$-MOCK there are no criteria to prevent the 'optimizing' of the partitions. This general view demonstrates conditions regarding the choice of the objective functions, and we presented a scenario where optimization is not required, answering our second research question.

\section{Conclusion}\label{sec:remarks}

This paper proposed and presented an analysis of the admissibility of clustering criteria, in support to defininig objective functions in evolutionary multi-objective clustering approaches.  
Our experiments demonstrated a lack of objective functions that can be widely admissible in different datasets, making the development of EMOCs for generalized clustering difficult. 

Furthermore, we highlighted the importance of aligning the choice of the objective function and the initialization strategy in designing the EMOCs.
In general, the use of a traditional clustering algorithm in the initialization provides solutions that reach the boundaries of the search space in terms of some criteria. Thus, optimizing the objective functions that consider such criteria is not required, thus other complementary criteria should be applied in the optimization. 
 In constrast, the criteria applied in the initialization can be taken as ``restrictions'', to determine the feasible search region in EMOCs. It is important to note that, in general, the EMOCs do not use the explicit restrictions (see Eq.~\ref{mop_r1} in Section \ref{section:background}). In many cases, the ``restrictions'' are represented as objective functions without prior notice, which could lead to a mistake regarding the understanding of which objectives are optimized. 
Thus, our study helps the understanding of the concept of admissibility to support the better choice of the objective functions, considering the different roles that the objective function can perform in the evolutionary multi-objective optimization.

For future work, we consider that real-life datasets should be analyzed to expand the admissibility analysis of the objective functions. In this paper, we only used artificial datasets to make it possible to analyze the data structures in the datasets and correlate them with the clustering criteria and the initialization strategy. Moreover, specific clustering criteria regarding application domain features can be explored in the analysis of real-life datasets.    

Another interesting direction of research is the improvement of the EMOC design to determine whether the objective functions should be optimized for different datasets or even the definition of new objective functions that could be widely admissible in different datasets.



\section*{Acknowledgement}
This work was partially supported by the National Council for Scientific and Technological Development (CNPq), Brazil.

\bibliographystyle{elsarticle-num}
\bibliography{ref}








\appendix

\begin{landscape}
\section{The box-plot of the ARI obtained in each initialization algorithm.}\label{app1}
\setcounter{figure}{0}
The Fig. \ref{fig:boxplotA1} presents the box-plot of the ARI considering the different initialization algorithms for each dataset. 

\begin{figure}[!htb]
	\centering
  \subfloat[\texttt{R15}]{\includegraphics[width=0.16\linewidth, height=0.3\textheight]{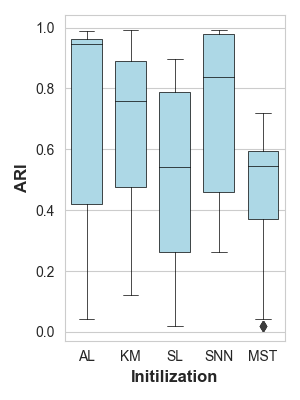}\label{R15plot}}
	 	\hspace{0.05mm}
  \subfloat[\texttt{D31}]{\includegraphics[width=0.16\linewidth, height=0.3\textheight]{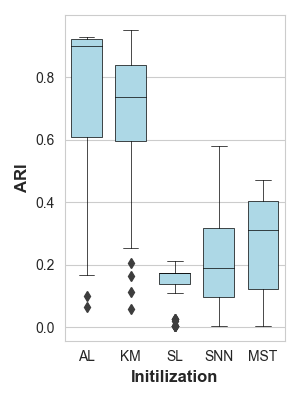}\label{D31plot}}
	 	\hspace{0.05mm}
	\subfloat[\texttt{Engytime}]{\includegraphics[width=0.16\linewidth, height=0.3\textheight]{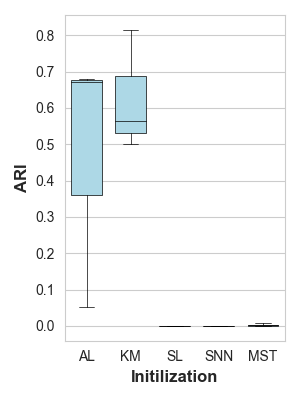}\label{Engytimeplot}}
 	\hspace{0.05mm}
	 \subfloat[\texttt{Sizes5}]{\includegraphics[width=0.16\linewidth, height=0.3\textheight]{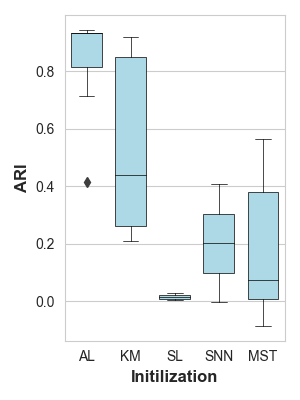}\label{Sizes5plot}}
	 	\hspace{0.05mm}
	\subfloat[\texttt{Square1}]{\includegraphics[width=0.16\linewidth, height=0.3\textheight]{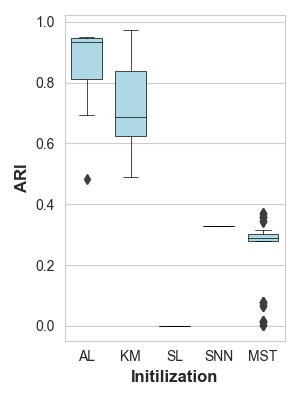}\label{Square1plot}}
	 	\hspace{0.05mm}
  \subfloat[\texttt{Square4}]{\includegraphics[width=0.16\linewidth, height=0.3\textheight]{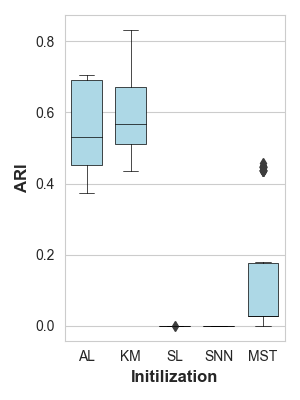}\label{Square4plot}}
\end{figure}

\begin{figure}[!htb]
\centering
	\ContinuedFloat 
	\subfloat[\texttt{Twenty}]{\includegraphics[width=0.16\linewidth, height=0.3\textheight]{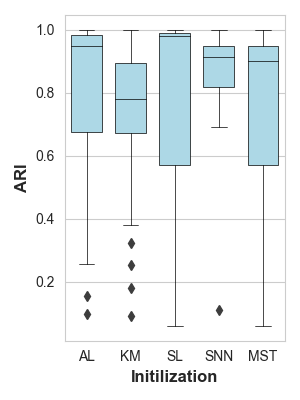}\label{Twentyplot}}
	 \hspace{0.05mm}
  \subfloat[\texttt{Fourty}]{\includegraphics[width=0.16\linewidth, height=0.3\textheight]{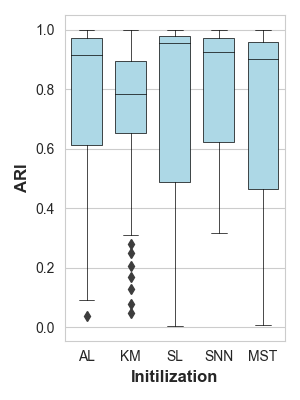}\label{Fourtyplot}}
	 \hspace{0.05mm}
	\subfloat[\texttt{Sph\_5\_2}]{\includegraphics[width=0.16\linewidth, height=0.3\textheight]{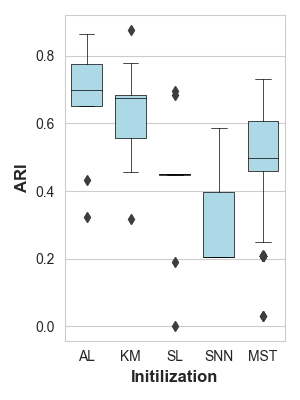}\label{Sph52plot}}
	\hspace{0.05mm}
  \subfloat[\texttt{Sph\_6\_2}]{\includegraphics[width=0.16\linewidth, height=0.3\textheight]{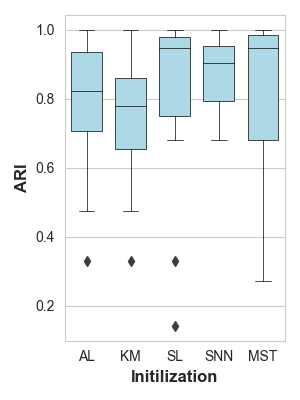}\label{Sph62plot}}
	 \hspace{0.05mm}
	\subfloat[\texttt{Sph\_9\_2}]{\includegraphics[width=0.16\linewidth, height=0.3\textheight]{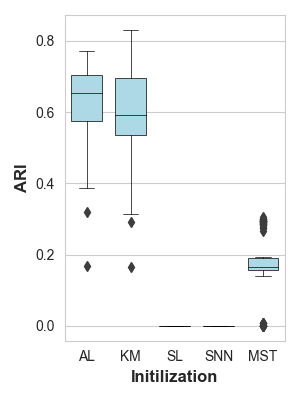}\label{Sph92plot}}
	 \hspace{0.05mm}
  \subfloat[\texttt{Sph\_10\_2}]{\includegraphics[width=0.16\linewidth,height=0.3\textheight]{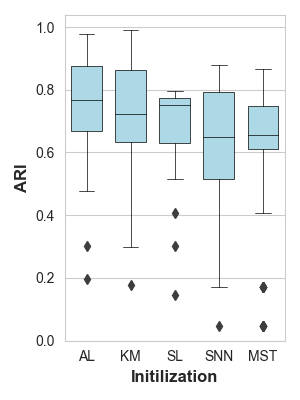}\label{Sph102plot}}
	\end{figure} 

\begin{figure}[!htb]
\centering
	\ContinuedFloat 
	\subfloat[\texttt{ds2c2sc13\_V1}]{\includegraphics[width=0.16\linewidth, height=0.3\textheight]{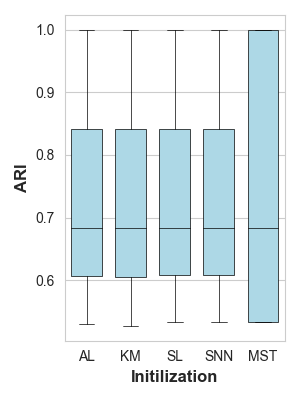}\label{ds2c2sc13V1plot}}
	 \hspace{0.05mm}
 \subfloat[\texttt{ds2c2sc13\_V2}]{\includegraphics[width=0.16\linewidth, height=0.3\textheight]{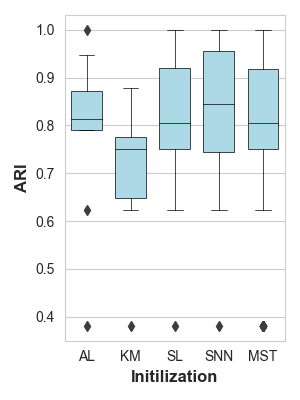}\label{ds2c2sc13V2plot}}
  \hspace{0.05mm}
 \subfloat[\texttt{ds2c2sc13\_V3}]{\includegraphics[width=0.16\linewidth, height=0.3\textheight]{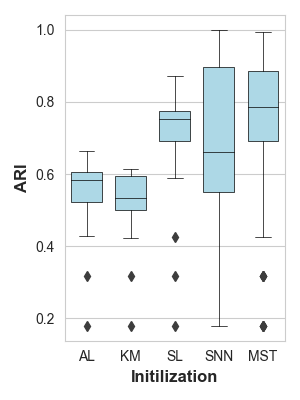}\label{ds2c2sc13V3plot}}
   \hspace{0.05mm}
	\subfloat[\texttt{Long1}]{\includegraphics[width=0.16\linewidth, height=0.3\textheight]{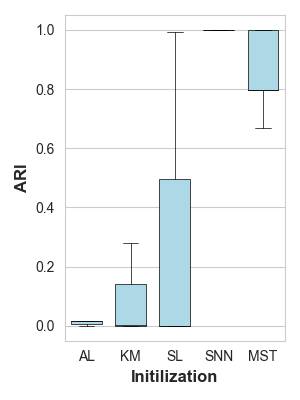}\label{Long1plot}}
	 \hspace{0.05mm}
	\subfloat[\texttt{Pat2}]{\includegraphics[width=0.16\linewidth, height=0.3\textheight]{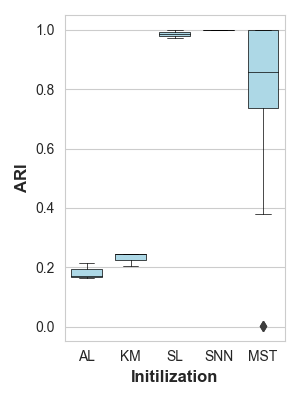}\label{Pat2plot}}
	 \hspace{0.05mm}
  \subfloat[\texttt{Spiral}]{\includegraphics[width=0.16\linewidth, height=0.3\textheight]{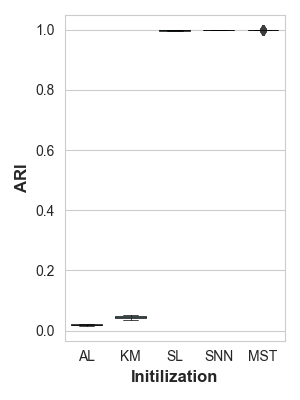}\label{Spiralplot}}
	\end{figure}

\begin{figure}[!htb]
\centering
	\ContinuedFloat 
	\subfloat[\texttt{3MC}]{\includegraphics[width=0.16\linewidth, height=0.3\textheight]{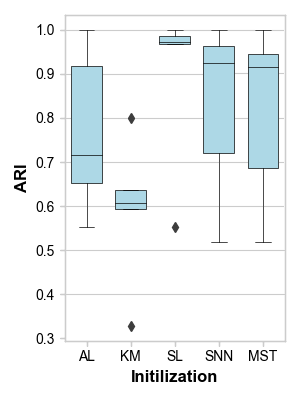}\label{3MCplot}}
	 \hspace{0.05mm}
  \subfloat[\texttt{DS-850}]{\includegraphics[width=0.16\linewidth, height=0.3\textheight]{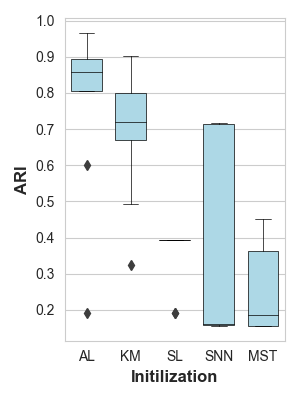}\label{DS850plot}}
  \hspace{0.05mm}
  \subfloat[\texttt{Aggregation}]{\includegraphics[width=0.16\linewidth, height=0.3\textheight]{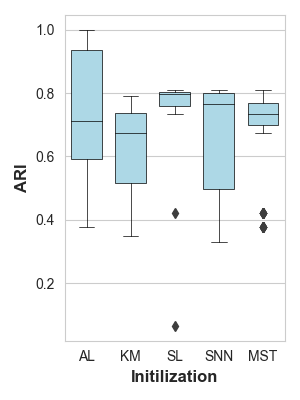}\label{Aggregationplot}}
  \hspace{0.05mm}
	\subfloat[\texttt{Complex9}]{\includegraphics[width=0.16\linewidth, height=0.3\textheight]{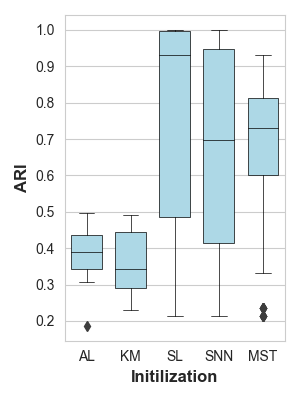}\label{Complexplot}}
	 \hspace{0.05mm}
	\subfloat[\texttt{Pat1}]{\includegraphics[width=0.16\linewidth, height=0.3\textheight]{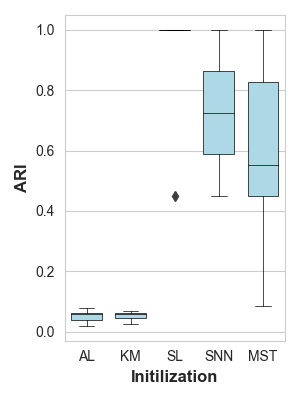}\label{Pat1plot}}
	 \hspace{0.05mm}
  \subfloat[\texttt{Spiralsquare}]{\includegraphics[width=0.16\linewidth, height=0.3\textheight]{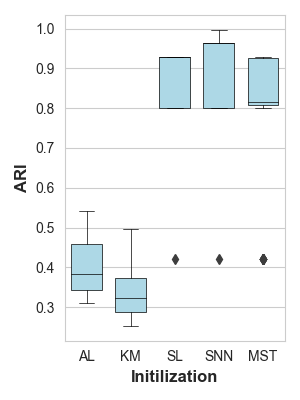}\label{Spiralsquareplot}}
\caption{{Box-plot of the ARI considering different initialization algorithms for each dataset}}
   \label{fig:boxplotA1}
\end{figure}
 \end{landscape}

 \begin{landscape}
\section{The box-plot of the ARI for the results generated by $\Delta$-MOCK.}\label{app2}
\setcounter{figure}{0}
The Fig. \ref{fig:boxplotA2} presents the box-plot of the ARI considering different objective function for each dataset.

\begin{figure}[!htb]
\centering
  \subfloat[\texttt{R15}]{\includegraphics[width=0.16\linewidth, height=0.3\textheight]{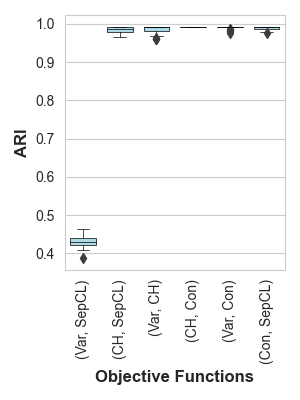}\label{R15plotf}}
	 \hspace{0.05mm}
  \subfloat[\texttt{D31}]{\includegraphics[width=0.16\linewidth, height=0.3\textheight]{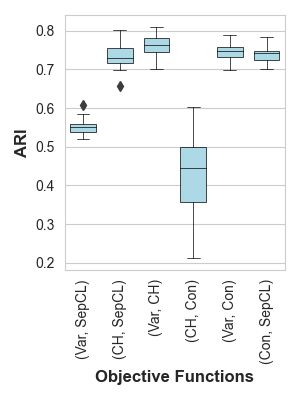}\label{D31plotf}}
   \hspace{0.05mm}
	\subfloat[\texttt{Engytime}]{\includegraphics[width=0.16\linewidth, height=0.3\textheight]{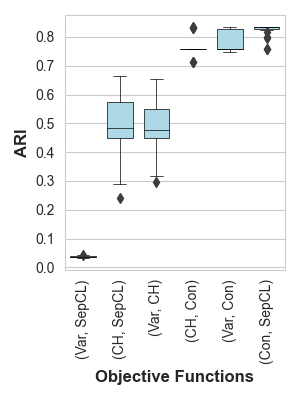}\label{Engytimeplotf}}
   \hspace{0.05mm}
	\subfloat[\texttt{Sizes5}]{\includegraphics[width=0.16\linewidth, height=0.3\textheight]{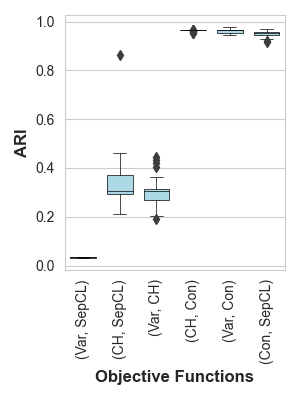}\label{Sizes5plotf}}
	 \hspace{0.05mm}
	\subfloat[\texttt{Square1}]{\includegraphics[width=0.16\linewidth, height=0.3\textheight]{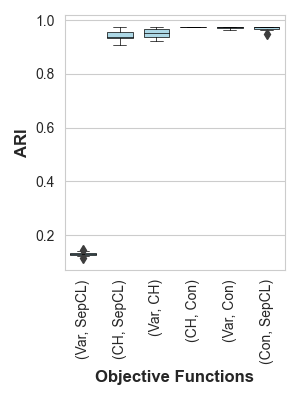}\label{Square1plotf}}
	 \hspace{0.05mm}
  \subfloat[\texttt{Square4}]{\includegraphics[width=0.16\linewidth, height=0.3\textheight]{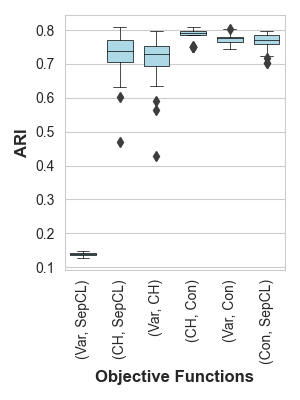}\label{Square4plotf}}
	\end{figure}
	
\begin{figure}[!htb]
	\ContinuedFloat 
	\centering
	\subfloat[\texttt{Twenty}]{\includegraphics[width=0.16\linewidth, height=0.3\textheight]{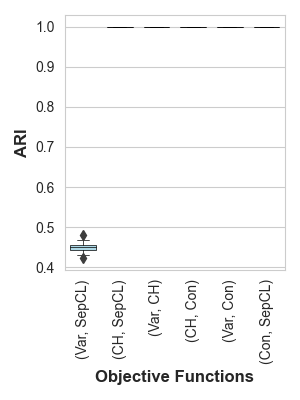}\label{Twentyplotf}}
	 \hspace{0.05mm}
  \subfloat[\texttt{Fourty}]{\includegraphics[width=0.16\linewidth, height=0.3\textheight]{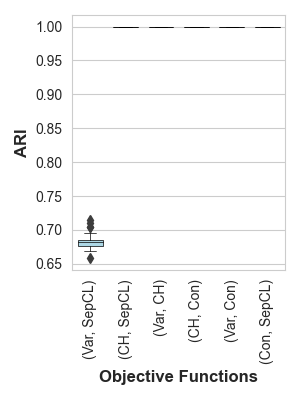}\label{Fourtyplotf}}
	 \hspace{0.05mm}
	\subfloat[\texttt{Sph\_5\_2}]{\includegraphics[width=0.16\linewidth, height=0.3\textheight]{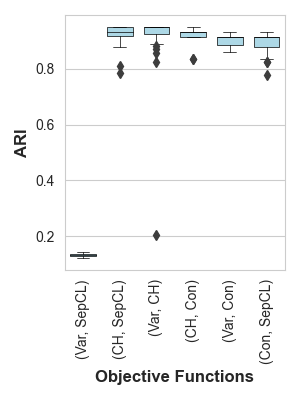}\label{Sph52plotf}}
	 \hspace{0.05mm}
	\subfloat[\texttt{Sph\_6\_2}]{\includegraphics[width=0.16\linewidth, height=0.3\textheight]{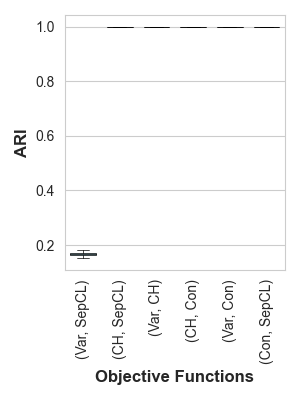}\label{Sph62plotf}}
	 \hspace{0.05mm}
	\subfloat[\texttt{Sph\_9\_2}]{\includegraphics[width=0.16\linewidth, height=0.3\textheight]{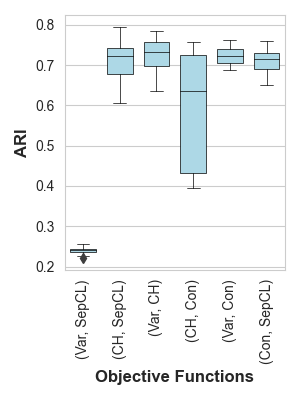}\label{Sph92plotf}}
	 \hspace{0.05mm}
  \subfloat[\texttt{Sph\_10\_2}]{\includegraphics[width=0.16\linewidth, height=0.3\textheight]{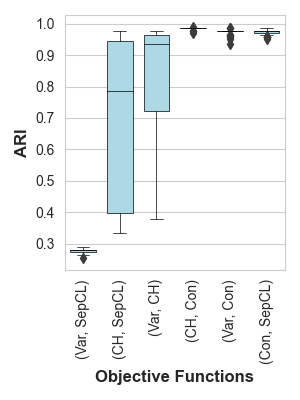}\label{Sph102plotf}}
\end{figure}

\begin{figure}[!htb]
	\ContinuedFloat 
	\centering
	\subfloat[\texttt{ds2c2sc13\_V1}]{\includegraphics[width=0.16\linewidth, height=0.3\textheight]{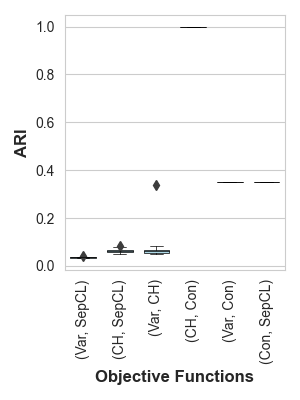}\label{ds2c2sc13V1plotf}}
	 \hspace{0.05mm}
 \subfloat[\texttt{ds2c2sc13\_V2}]{\includegraphics[width=0.16\linewidth, height=0.3\textheight]{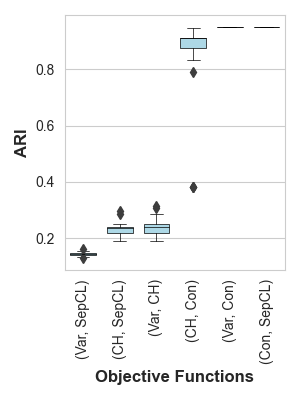}\label{ds2c2sc13V2plotf}}
  \hspace{0.05mm}
 \subfloat[\texttt{ds2c2sc13\_V3}]{\includegraphics[width=0.16\linewidth, height=0.3\textheight]{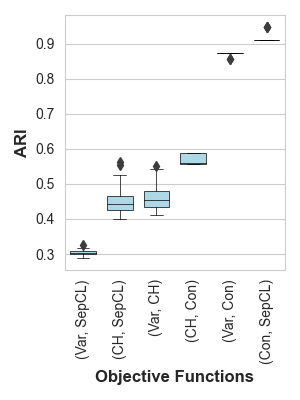}\label{ds2c2sc13V3plotf}}
  \hspace{0.05mm}
	\subfloat[\texttt{Long1}]{\includegraphics[width=0.16\linewidth, height=0.3\textheight]{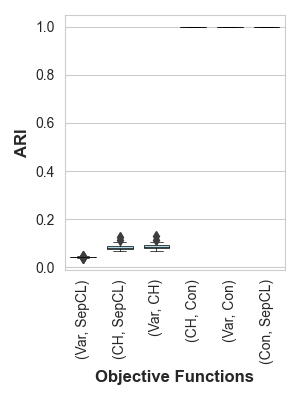}\label{Long1plotf}}
	 \hspace{0.05mm}
	\subfloat[\texttt{Pat2}]{\includegraphics[width=0.16\linewidth, height=0.3\textheight]{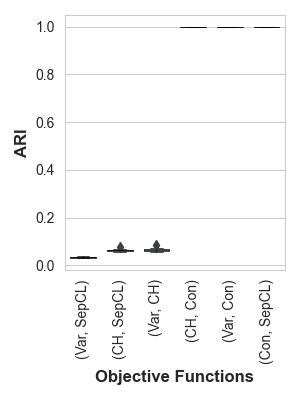}\label{Pat2plotf}}
	 \hspace{0.05mm}
  \subfloat[\texttt{Spiral}]{\includegraphics[width=0.16\linewidth, height=0.3\textheight]{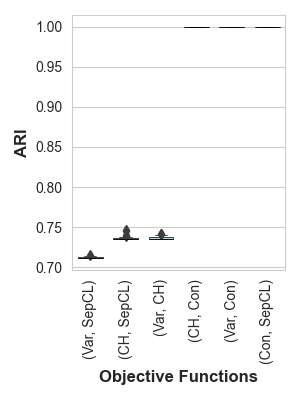}\label{Spiralplotf}}
\end{figure}

\begin{figure}[!htb]
	\ContinuedFloat 
	\centering
	\subfloat[\texttt{3MC}]{\includegraphics[width=0.16\linewidth, height=0.3\textheight]{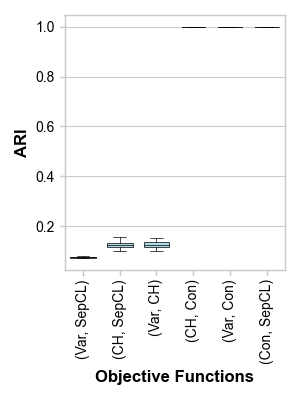}\label{3MCplotf}}
	 \hspace{0.05mm}
  \subfloat[\texttt{DS-850}]{\includegraphics[width=0.16\linewidth, height=0.3\textheight]{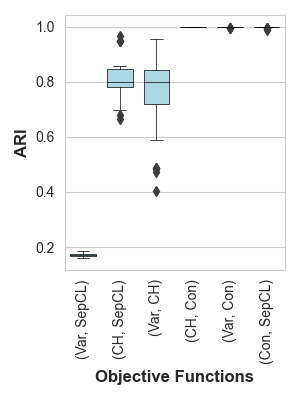}\label{DS850plotf}}
	 \hspace{0.05mm}
	\subfloat[\texttt{Aggregation}]{\includegraphics[width=0.16\linewidth, height=0.3\textheight]{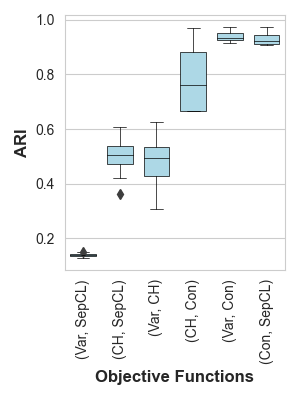}\label{Aggregationplotf}}
	 \hspace{0.05mm}
	\subfloat[\texttt{Complex9}]{\includegraphics[width=0.16\linewidth, height=0.3\textheight]{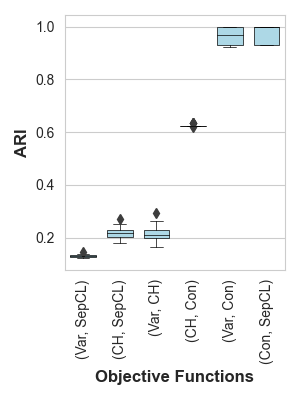}\label{Complexplotf}}
	 \hspace{0.05mm}
	\subfloat[\texttt{Pat1}]{\includegraphics[width=0.16\linewidth, height=0.3\textheight]{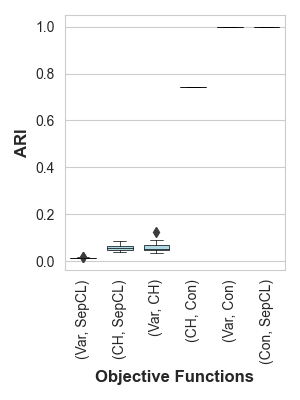}\label{Pat1plotf}}
	 \hspace{0.05mm}
  \subfloat[\texttt{Spiralsquare}]{\includegraphics[width=0.16\linewidth, height=0.3\textheight]{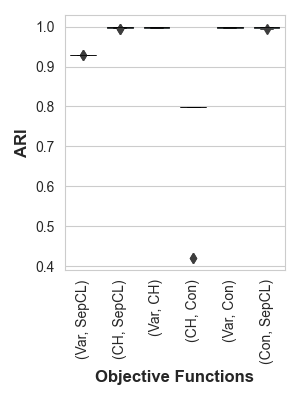}\label{Spiralsquareplotf}}
	\caption{{Box-plot of the ARI considering different objective functions for each dataset}}
   \label{fig:boxplotA2}
\end{figure}

 \end{landscape}
\end{document}